\documentclass[a4paper,11pt]{report}
\pdfoutput=1
\usepackage{float}
\usepackage[table,xcdraw]{xcolor}
\usepackage{subcaption}
\usepackage{tikz}
\usetikzlibrary{arrows}
\usepackage{rotating}
\usepackage{booktabs}
\usepackage{multirow}
\usepackage{tabulary}
\usepackage{graphicx}
\graphicspath{{./figures/}}
\usepackage{float}
\usepackage{algorithmic}
\usepackage[ruled,vlined,nofillcomment,titlenumbered]{algorithm2e}
\usepackage{verbatim}
\usepackage{latexsym}
\usepackage{mathchars}
\usepackage{mathtools}
\usepackage{amssymb}
\usepackage{setspace}
\usepackage{chngpage,calc}
\usepackage{color} %
\usepackage[toc,page]{appendix}

\usepackage{titlesec}
\titlespacing\subsection{0pt}{5pt plus 4pt minus 0pt}{-5pt plus 2pt minus 0pt}
\titlespacing\subsubsection{0pt}{5pt plus 4pt minus 0pt}{-5pt plus 2pt minus 0pt}
\SetKwFor{ON}{on}{do}{}
\SetKwFor{REPEAT}{repeat}{}{}
\SetKwFor{CLASS}{class}{}{}
\SetKwFor{METHOD}{method}{}{}
\SetKwFor{PROCEDURE}{procedure}{}{}
\SetKwFor{FUNCTION}{function}{}{}
\SetKw{AND}{and}
\SetKw{OR}{or}
\SetKw{NOT}{not}
\SetKw{TO}{to}
\SetKw{TRUE}{true}
\SetKw{FALSE}{false}
\SetKw{DOWNTO}{downto}
\SetKw{SEND}{send}
\SetKw{RETURN}{return}

\newcommand\restr[2]{{\left.\kern-\nulldelimiterspace #1\vphantom{\big|}\right|_{#2}}}

\newcommand{\myCaption}[2]{\caption[#1]{\textbf{#1}. \textit{#2}}}

\newcommand{\Real}{\bbbr}
\newcommand{\real}{\Real}

\newcommand{\NN}{{\sf I\kern-0.14emN}}   % Natural numbers
\newcommand{\ZZ}{{\sf Z\kern-0.45emZ}}   % Integers
\newcommand{\QQQ}{{\sf C\kern-0.48emQ}}   % Rational numbers
\newcommand{\RR}{{\sf I\kern-0.14emR}}   % Real numbers

\newcommand{\GG}{{\bf G}}

\newcommand{\XX}{{\mathcal{X}}}

\newcommand{\cc}{{\bf c}}

\newcommand{\ww}{{\bf w}}
\newcommand{\xx}{{\bf x}}

\newcommand{\aalpha}{{\boldsymbol{\alpha}}}

\newcommand{\imply}{\Rightarrow}
\newcommand{\bimply}{\Leftrightarrow}

\newcommand{\normallinespacing}{\renewcommand{\baselinestretch}{1.5} \normalsize}

\newcommand{\syncc}{~\stackrel{\textstyle \rhd\kern-0.57em\lhd}{\scriptstyle L}~}

\newtheorem{definition}{Definition}[chapter]

\newcommand{\dualFormulation}{
\begin{equation}
\begin{split}
\underset{\boldsymbol{\lambda}\in \real^n}{\textrm{max}} &\hspace{10pt}\sum_{i=1}^{n}\lambda_i - \frac{1}{2}\sum_{i=1}^{n}\sum_{j=1}^{n}\lambda_i\lambda_j \hspace{2pt} y_i y_j \hspace{2pt} \Phi(\xx_i)^T \Phi(\xx_j) \\
\textrm{subject to:}&\hspace{10pt}\lambda_i\geq 0 \hspace{15pt} \textrm{for all }i \in \{1,...,n\}\\
&\hspace{10pt}\sum_{i=1}^{n}\lambda_i y_i = 0
\end{split}
\end{equation}
}

\newcommand{\dualKernelFormulation}{
\begin{equation}
\begin{split}
\underset{\boldsymbol{\lambda}\in \real^n}{\textrm{max}} &\hspace{10pt}\sum_{i=1}^{n}\lambda_i - \frac{1}{2}\sum_{i=1}^{n}\sum_{j=1}^{n}\lambda_i\lambda_j \hspace{2pt} y_i y_j \hspace{2pt} k(\xx_i,\xx_j)\\
\textrm{subject to:}&\hspace{10pt}\lambda_i\geq 0 \hspace{15pt} \textrm{for all }i \in \{1,...,n\}\\
&\hspace{10pt}\sum_{i=1}^{n}\lambda_i y_i = 0
\end{split}
\end{equation}
}

\begin{document}

\title{Multiple protein feature prediction with statistical relational learning}

\author{Luca Masera}

\normallinespacing
\maketitle

\chapter{Introduction}

\section{Motivations}

Proteins are the workhorse molecules of life, they take part in almost everything what happens in and outside the cells, from building structures such as hair or nails, to enzymes, which accelerate chemical reactions. It is sufficient to consider some number to understand their importance and variety: the total number of proteins in human cells is estimated to be between $250,000$ to one million and the dry weight of our bodies is made by about the $75\%$ of proteins \cite{peso}. Hence, it is easy to understand that an accurate analysis of their features and properties is crucial to understand the deepest mechanism of life.

Since their identification in the eighteenth century, proteins have been a focal point of micro-biologist researches, but it is just after the half of the twentieth century that the study of proteins makes its biggest progresses. Indeed, in that years Sanger accurately determined the correct amino acid sequence of insulin, and Max Perutz and Sir John Cowdery Kendrew respectively resolved the three dimensional structure of hemoglobin and myoglobin through X-ray crystallography. These discoveries opened the floodgates to a completely new set of protein analysis techniques. In the 80's indeed, biologist recognized the potential of the firsts portable computer and began to store their newly obtained protein sequences and structures data in digital format. Since them marriage between computer science and biology has been long and fruitful. 

Protein sequencing however proceeds much faster than the ability of wet biologists to annotate the genes product. Moreover computer science technological breakthroughs and next generation sequencing techniques exacerbate this difference. Indeed, at the beginning of 2015 there are almost 200 millions\footnote{http://www.ncbi.nlm.nih.gov/genbank/statistics} genes sequenced and only the $0.25\%$\footnote{http://web.expasy.org/docs/relnotes/relstat.html} of them have been manually curated and annotated. These statistics clarify that new and reliable computational approaches are needed to fill this gap and in this field machine learning techniques find breeding ground for their application to biological problems. In fact in the last decades they have been widely applied to the protein feature prediction task, but their results are yet not precise enough to guarantee reliable prediction to biologists. This research area is therefore nowadays very hot and alive.

\section{Contributions}
The aim of this work is to face the protein feature prediction, both from a statistical and a relational point of view. This bioinformatics task is a multi-labeled problem where it is needed to assign multiple annotations to a single protein. However this labels are not completely independent one from the other and therefore knowledge on the relations between them could positively impact on the quality of the predictions.

In standard machine learning algorithms indeed, knowledge lies only into the dataset and the usual way to enrich it with field-specific background information is via sophisticated kernel functions. Unfortunately this operation is not always effective, it is indeed hard to formulate relations between learned classes. Semantic-Based Regularization (SBR) \cite{SBRS}, a state-of-the-art relation statistical learning framework developed in the University of Siena, provides a possible solution to this lack allowing the supervisor to instill background information into the well established kernel machine learning process in form of first order logic constraints.

SBR gives us the tools to integrate both aspects and achieve better protein feature prediction. In particular, we focus our attention primarily on the prediction of the Gene Ontology annotation. 

From the relational point of view the aim was to develop first order logic constraints that express knowledge on the relationships between the predicted labels. The first set of logical rules has been obtained from the consistency property of Gene Ontology, i.e. if an annotation is assigned to a gene product (in our case proteins), then all the annotation ancestors in the hierarchy must be associated with that gene product. The second set of information exploited in this work comes from a protein-protein interaction dataset, whence we derived additional rules to submit to our framework.

In addition to the effort done in generating valid and informative rules that could effectively improve the learning process, we implemented four kernel functions that exploited different aspect of the protein, going from the simple amino acidic sequence to the expression levels of the genes responsible for their creation.

\section{Structure of the document}

\subsubsection{Background}
Our study is about a very specific field, i.e. prediction of protein function, so in the first part background knowledge are presented and defined to give the reader the basic tools to understand  the subject treated in this document.
At the beginning we will illustrate the importance of proteins, which play a crucial role in our work, describing their structure and highlighting the relevance of these bio-molecules for life. 
Subsequently the main bioinformatics resources for retrieving (e.g. protein databases and gene ontology) and analyzing (e.g. sequence alignment algorithms) protein information are provided. Moreover, it is taken an overview of the literature about the several ways how this yet hard task has been faced until now.
The considerations made in these sections will also be useful to understand the choices made in the kernel functions selection and the biological conclusions in the results.

After the biological excursus, the machine learning tools used in this work are presented and illustrated. After the explanation of the features and theoretical firmness of kernel machines, the document comes to the description of SBR, the relational statistical learning framework exploited in the protein feature prediction task.

\subsubsection{Methods}
The purpose of this chapter is to explain the method developed in the experimental phase of the work. 
After the formalization of the our task, it is provided the description of the kernel functions used, which will have an important part in the results discussion, before proposing the developed rules to inject prior knowledge into the learning process. The methods chapter ends with the description of the pipeline implemented to automatize the generation and execution of experiments and the metrics used to evaluate them. 

\subsubsection{Results}
Since this work is an experimental one, this chapter covers the most important part of the document, explaining the results obtained with our relational statistical learning setting on more then 1500 \emph{S. cerevisiae}'s proteins. The discussion is divided based on the type of feature analyzed and will be focused on the effects of kernel functions and set of rules used.

\subsubsection{Conclusion}
The conclusions of this document are drawn in this final chapter, in which  the goals that we achieved are presented and explained, correlated to the baseline purposes of our work and the novelties found during the developing of it. It will be specially focused on the confirmed and disproved hypothesis that we set at the very first beginning. Moreover will be also found useful ideas for future works on this topic, that could further been explored.

\chapter{Biological background}

\label{ch:background}

\section{Proteins}
This work is mostly centered around the prediction of protein features. It is therefore important to give the reader an overview on the main concept related to the argument and the taste of the importance of the role that proteins play for life.

Proteins are biological macromolecules composed of one or more amino acids chains and, together with lipids, sugars and nucleic acids, are considered to be the fundamental building blocks of life. Examples of important biological roles played by proteins may include catalytic activity (e.g. amylase contribute to digestion), signaling and regulation (e.g. somatropin also known as growth hormone stimulates growth and cell reproduction in humans and other animals), molecule transport (e.g. Na$^+$/K$^+$-ATPase allows the exchange of sodium and potassium in neural cells), structural support (e.g. tubulin, the monomer of the microtubules in the cytoskeleton) and movement (e.g.  myosin allows muscle contraction).

The biological information needed for generating new copies of proteins is contained in the genes, that are DNA portions. The process by which a gene is copied into a RNA filament is called transcription. The transcripted RNA, called messenger RNA (mRNA), is then synthesized into a amino-acid chain by the ribosome, which is itself a protein (with some RNA chains). This step is called translation because each triplet of nucleotides in the mRNA is translated into one of the 22 amino acids and linked together to form a new protein. Transcription and translation are crucial for life, so much that together with the DNA replication, which is also mediated by proteins,  form the central dogma of molecular biology (Figure \ref{fig:central}).

\begin{figure}
\centering
\begin{tikzpicture}[->,>=stealth',shorten >=1pt,auto,node distance=4.5cm,
  main node/.style={rectangle,draw,font=\large\bfseries}]

  \node[main node] (1) {DNA};
  \node[main node] (2) [right of=1] {RNA};
  \node[main node] (3) [right of=2] {Protein};

  \path[every node/.style={font=\large}]
    (1) edge [loop left] node {Replication} (1)
    	edge node {Transcription} (2)
    (2) edge node {Translation} (3);
\end{tikzpicture}
\myCaption{Central Dogma of Molecular Biology}{The Central Dogma of Molecular Biology describes the three fundamental processes for life. Is interesting to notice that all these are carried out by proteins.}
\label{fig:central}
\end{figure}

\subsection{Structure}
Proteins are long polymers and amino acids are their monomers. Amino acids are organic molecules composed of a common backbone, consisting of amine (-NH$_2$) and carboxylic acid (-COOH) functional groups, along with a specific side-chain, that characterize each amino acid. Indeed, the physico-chemical properties of the side chain determine the structure and the shape of the whole protein.

After the translation, proteins fold themselves in complex three dimensional structures, which will widely determine their functionalities. During the folding process, not only the amino acids composition is important, but also the environment plays a critical role. Indeed, according to it, some amino acids may “prefer” to face the outside of the protein or the core. If the folding does not succeed the protein is useless or, in some cases, even dangerous. It is indeed known that many degenerative diseases like the Alzheimer, the Cystic fibrosis or the BSE (commonly known as "mad cow" disease) are caused by the misfolding of proteins. Therefore, there exist specialized proteins (called molecular chaperones) and biological process (translocation) that try to guarantee to proteins the right folding environment.

\begin{figure}
\centering
\includegraphics[width=1\textwidth]{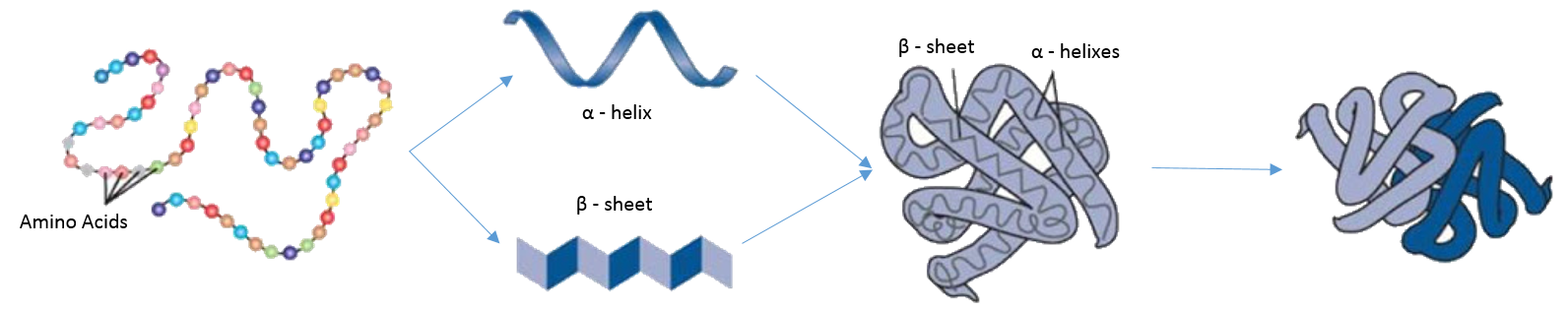}
\myCaption{Protein structure}{In figure are shown the schematic representation of the four levels of the protein structure.}
\label{fig:folding}
\end{figure}

The structure of a protein is usually analyzed at four different levels (Figure \ref{fig:folding}), which are: 
\begin{itemize}
\item Primary structure: is the lowest of the four levels and is determined by the bare sequence of amino acids. Nowadays it is reasonable easy to obtain this information by translating the nucleotide triplets in coding sequences of genes obtained by genome sequencing, but is less informative than the higher levels.

\item Secondary structure: represents how the peptidic backbone interacts with itself through hydrogen bonds. At this level specific folding shapes can be identified, i.e. the alpha helix and the beta sheet. This firs level of folding occurs just after the translation (few milliseconds) and is mainly caused by hydrophobic behaviors of the amino acids.

\item Tertiary structure: describes how the secondary structures interact with themselves forming the three dimensional shape. The structure is stabilized through hydrogen bonds and disulfide bridges (covalent bonds created post-translation). Information relatively about this levels are still hard and expensive to obtain. Bioinformatic methods are not powerful enough to face this problem, so techniques like NMR (Nuclear Magnetic Resonance) or X-ray crystallography have to be applied.

\item Quaternary structure: some proteins are constituted by more than one peptidic chain or are part of protein complexes. These interactions are described by the quaternary structure. Studying this structure level is usually even harder than for the tertiary one. Due to the steric size of protein complexes, they can not be easily crystallized and analyzed with X-ray crystallography. In this cases in then used the cryo-electron microscopy technique that allows to literally freeze the structure of a protein and investigate it through an electronic microscopy.
\end{itemize}

Even after having finished the folding process, the structure of a protein is not rigid and immutable. Their three dimensional structure can indeed change in time depending on the surrounding environment (temperature, pH, voltage, ion concentration, phosphorylation or ligand binding), which can actually trigger their functionalities. It is, for example, the case of myosin which interacts with ATP molecule. ATP molecules are very energetic and their hydrolysis induces a conformational change in the protein that, as a final result, allows the muscle contraction.

Evolution has deeply shaped living beings, giving them very different forms and features, from their appearance to their molecular level. Proteins are no exception. Despite this there are portion of them, which are strongly conserved between different proteins of the same organism but also between species. This pieces of proteins are called domains and are usually characterized by a compact structure and a specific folding. Protein domains are strongly related to their function and are combined in proteins like building blocks. Their conservation can be used as key for identifying known domains in proteins according to their primary sequence.

\section{Bioinformatic resources for proteins}

In the first years of the 80's two technological breakthroughs were gaining a foothold in the scientific word: the Sanger sequencing methods and the first computers. The former made it possible to acquire rapidly and accurately (for the time) the sequence of long stretches of DNA, the latter to store and analyze this information. Since then the marriage between biology and computer science has been long and successful. Many specific algorithms and resources have been developed by computer scientists and then applied by computational biologists to explore the deepest aspects of life. In this section are proposed the most relevant results of this fruitful marriage with a special focus on the protein analysis.

\subsection{Algorithms}
Because there are only 22 amino acids\footnote{Selenocysteine and Pyrrolysine are often not included in the list because very rare and until a few years ago considered of secondary importance.} that constitute proteins, proteins primary structure is usually represented as a string using an alphabet of 22 symbols  $\Sigma= \{$A, C, D, E, F, G, H, I, K, L, M, N, P, Q, R, S, T, U, V, W, Y$\}$. This makes proteins very suitable to be treated with string algorithm to detect similarities that can be related to functional or structural relationships.

\begin{figure}[b]
\centering
\small
\texttt{
>gi|129295|sp|P01013|OVAX\_CHICK GENE X PROTEIN (OVALBUMIN-RELATED){\color{white}LAPAE}	QIKDLLVSSSTDLDTTLVLVNAIYFKGMWKTAFNAEDTREMPFHVTKQESKPVQMMCMNNSFNVATLPAE
KMKILELPFASGDLSMLVLLPDEVSDLERIEKTINFEKLTEWTNPNTMEKRRVKVYLPQMKIEEKYNLTS
VLMALGMTDLFIPSANLTGISSAESLKISQAVHGAFMELSEDGIEMAGSTGVIEDIKHSPESEQFRADHP
FLFLIKHNPTNTIVYFGRYWSP\color{white}ESLKISQAVHGAFMELSEDGIEMAGSTGVIEDIKHSPESEQFRADHP
}
\myCaption{Example of amino acidic sequence in FASTA format}{This is an example of a primary sequence retrievable in one of the many primary databases. The "\textgreater" symbol introduces the description line, a series of meta information used to identify the sequence. The following sequence is then formatted in such a way that each line of text is shorter than 80 characters in length.}
\end{figure}

When protein sequences are compared, scoring function that reflect biological or statistical relevance of a couple of amino acids are employed. So, instead of using a binary function (1 if considering the same amino acid, 0 otherwise), the similarity score is obtained by substitution matrices like PAM (Point Accepted Mutation) \cite{PAM} or BLOSUM (Blocks Substitution Matrix) \cite{BLOSUM}. The first one expresses the evolutionary acceptability of a specific point mutation, the second instead encodes the empirical probability of an amino acid. Scoring functions are also able to deal with gaps in sequences, assigning them specific penalties based on position and length.

Sequence alignment algorithms are a vast class of algorithms whose goal is to identify the "best" possible between two sequences. The first distinction that can be done is in the definition of "best" alignment. There are indeed two different approaches to the problem, i.e. the local and the global. Global methods give more importance the total number of matching into the sequences. This can obviously leads to a fragmented solution that do not take into account the modular nature of proteins. Differently, local alignment algorithms attempt to maximize the alignment piece-wise. This approach is very useful when comparing dissimilar sequences in order to identify conserved regions such as motifs or domains.
A further differentiation involves the number of sequence treated per-time by the algorithm. 

\paragraph{Pairwise alignment} methods are efficient algorithms that consider two sequences at a time and could be local or global. They are usually used to deal with similarity searches in databases or to extrapolate information about an unknown sequence from the characteristics of a known one. The three primary pairwise alignment methods are:
\begin{itemize}
\item Dot-matrix method: is an exhaustive algorithm that tries all possible alignments between two sequences. It builds a matrix $D$, where the entries represent the quality of the alignment near to that position. It can be formalized as follows: given a sliding window $w$, which represents the number of consecutive amino acids to average, and two sequences $s$ and $t$, the entry in position $i,j$ of the dot-matrix is equal to $D_{i,j}=\frac{1}{w}\sum_{k\in 0...w}S(s_{i+k},t_{i+k})$ where $S$ is the scoring function used (e.g. binary, BLOSUM62, PAM120, ...). The out-coming matrix can be easily plotted mapping the values of each entry into a gray scale (an example in Figure \ref{fig:dot_matrix}). If the data are not too noisy and the size of the sliding window is correctly selected, repeated portions of the sequences and gaps can be clearly observed. However, this method can be mainly appreciated graphically because it is hard to extract match summary statistics and match positions of two sequences.
\begin{figure}[H]
\centering
\includegraphics[width=.8\textwidth]{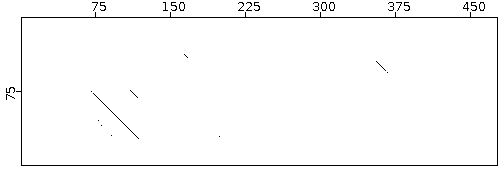}
\myCaption{Dot-plot of two {\it S. cerevisiae} proteins}{The plot shows the results of the dot matrix method on two \emph{S. cerevisiae} proteins, i.e. Q00776 and Q00381. A highly conserved region can be noticed. This corresponds indeed to a shared protein domain that is the AP complex, mu/sigma subunit.}
\label{fig:dot_matrix}
\end{figure}

\item Dynamic programming methods: are a class of algorithms that guarantees to find the optimum alignment, both in the local (Smith-Waterman algorithm) and in the global (Needleman-Wunsch algorithm) case. Given two sequences of size $n,m$ the two algorithms generate a $n\times m$ matrix that is progressively (dynamic programming) filled. Each entry represents the score of the best possible alignment up to that position. The difference between the two methods lies an the choice of the value to assign to the $i,j$ position, but, once the matrix is completed can reconstruct the final solution starting from the last entry in the matrix and going backwards up the first one. 

\item Word methods: they are heuristic methods that, differently from dynamic programming ones, are not guaranteed to reach a global optimum, but are significantly faster. The key idea of these class of methods is to use $k$-tuple (sequences of length $k$) to identify regions that will probably be in the final alignment, and from these starting the refinement. These methods often rely on substitution matrices for scoring pairs of amino acids and apply to gaps negative scores. The correct parameters of the matrix and of the size of $k$ may highly impact on the final results, but the choice is often more empirical than theoretical. The two most famous implementations of this class of algorithms are FASTA \cite{FASTA} and BLAST \cite{BLAST}, which are nowadays still widely used.

\end{itemize}

\emph{\null\hfill One or two homologous sequences whisper,\hspace{1cm} \newline\null\hfill a full multiple alignment shouts out loud.\hspace{1cm}\null}\vspace{-10pt}
\newline\newline\null\hfill(Hubbard et al.,1996)

\paragraph{Multiple alignment}
Multiple alignment (MSA) techniques attempt to align three or more sequences at the same time. These methods are often used to identify conserved region across a group of proteins highlighting phylogenetic relationships between them. They are, however, computationally very intensive\footnote{Most formulation of the multiple alignment problem lead to NP-complete combinatorial optimization problems \cite{multiple_seq_complex}} and heuristics are necessary for keeping them feasible.
\begin{itemize}
\item Dynamic programming methods: similarly to the pairwise case, dynamic programming can be applied to discover the optimal multiple alignment between $k$ sequences. However, in the MSA case, instead of building a two dimensional matrix, it requires a $k$-dimensional matrix, but the concept is analogous. The matrix is filled progressively and the score of each cell is computed according to the previously computed ones. Unfortunately the complexity is exponential in the number of sequences and therefore the problem becomes quickly intractable for increasing values of $k$.
\item Progressive alignment construction methods: are heuristic methods whose approach consists in two main steps. In the first one the algorithm generate a tree, called \emph{guide tree}, where two node are neighbor if they have the best pairwise alignment among all pairs. Then, starting from the couple of sequences with the highest similarity score, the final solution is obtained by incrementally adding new sequences according to the \emph{guide tree}. The most popular method of this class is the ClustalW algorithm \cite{CLUSTALW}.
\end{itemize}

\subsection{Databases}
The huge amount of biological data produced with the new sequencing methodologies has largely overtaken the biologist capability to analyze them manually, therefore new computational techniques are needed not only for mining informations, but also for storing and organizing them. In the last decades a vast variety of biological databases (BDB) has therefore appeared allowing easy storage and consultation of data. According to their content, databanks can be  classified as primary, the one storing primary sequences, and derived, which contain informations obtained by the analysis of primary sequences.

Important protein related BDB are:
\begin{itemize}
\item UniProt (The universal protein resource) \cite{UniProt}: is the biggest bioinformatics database created by the European Bioinformatic Institute (EBI), the Swiss Institute of Bioinformatics (SIB) and the Protein Information Resource (PIR). It collects protein sequences of most of living beings and viruses from the main publicly available databases and organize them in a comprehensive, non-redundant database (UniParc). Moreover part of this sequences are manually curated by experts. The annotation, such as protein function, subcellular localization, protein protein interaction domains and active sites are stored in the UniProtKB/Swiss-Prot database.
\item Protein Data Bank (PDB) \cite{PDB}: is the main repository for retrieving the three-dimensional structure of proteins and nucleic acids and is therefore a key resource of structural biology. Each entry in the database consists of all coordinates of all atoms in a specific protein. This data are submitted by biologists and biochemists from allover the world and typically obtained with X-ray crystallography or NMR spectrography. Stored information can be retrieved in a text format, but also as 3D browsable images. Since its foundation in 1973, the number of analyzed proteins is highly increased reaching at the beginning of 2015 almost 100000 proteins structures.
\item CATH \cite{greene2007cath,CATH}: is, similarly to SCOP, a hierarchical classification of protein domains. The name is itself the acronym of the four classification levels:
\begin{itemize}
\item Class: classifies the domains according to the their overall secondary structures composition (mainly alpha, mainly beta, alpha beta and few secondary structures).
\item Architecture: takes into account the general spatial arrangement of secondary structures of the domain.
\item Topology: is analogous to architecture, but including connectivity between secondary structures.
\item Homologous superfamily: this classification is manually curates and highlights an evolutionary relationship (at least two criteria from sequence, structure or function must be observed).
\end{itemize}
\item Structural Classification of Proteins (SCOP)\cite{SCOP}: is biological database containing a structural classification of protein domains. Its aim is to provide an evolutionary relations between proteins. At a first look the levels of SCOP (Class, Fold, Superfamily, Family etc) are similar to the CATH ones, but there are some classification differences.
\end{itemize}

\section{Gene Ontology}
\label{sec:GO}
The lack of a standard terminology in the biological field can lead to inefficient communications and ambiguous data sharing in the scientific community. Gene Ontology (GO) \cite{GO} is a bioinformatics project -part of `The OBO Foundry'- whose purpose is to fill this lack providing a controlled vocabulary of univocally defined terms representing gene product properties of all species.

\subsection{The structure}
Gene Ontology is a manually curated structured ontology with the aim to provide a unique and unequivocal description of all gene products. This information consist of an ID, a name, the GO domain and a natural language definition listing its main features. All the entries are organized in three direct acyclic graphs (DAGs), each one having as root a specific GO domain, which are:
\begin{itemize}
\item cellular component: the parts of a cell or its extra-cellular environment,
\item molecular function: the elemental activities of a gene product at the molecular level, such as binding or catalysis,
\item biological process: operations or sets of molecular events with a defined beginning and end, pertinent to the functioning of integrated living units: cells, tissues, organs and organisms.
\end{itemize}
In addition to the annotation, relation plays a very important role in the GO structure definition, giving fundamental information about relationships between the various terms.
\begin{itemize}
\item is a: is the main relation and forms the basic structure of GO. A GO term A is said to be in a \emph{is a} relation with a GO term B if A is a subtype of B. Therefore $A \imply B$, but not vice-versa. This relation is transitive, indeed taking the \emph{is a} chain of mitochondrion, intra-cellular organelle and organelle as example, it is clear that a mitochondrion is an organelle and not every organelle is mitochondrion.
\item part of: is similar to the \emph{is a} relation, but it express the concept of being part of something (cellular component) or take part in something (biological process and molecular function). As the \emph{is a} relation, \emph{part of} is transitive, indeed if A \emph{part of} B and B \emph{part of} C, then A \emph{part of} C. This relation can be trans-hierarchy, e.g. a Molecular Function can be \emph{part of} a Biological Process.
\item regulates: this relation represents the ability of a GO term to directly affects the manifestation of another. It can be further specified if the regulation is positive (at the growth of the first the second grows) or negatively (at the growth of the first the second decreases).
\item occurs in: this relation expresses a locational relationship between a BP and a CC term. It is probably the most informative trans-hierarchy relation, but unfortunately it occurs rarely in the GO dags.
\end{itemize}

A fundamental property considered in this work, which computational models often ignore and also neglect, is what in \cite{TPR} is called the True Positive Rule (TPR). This property derives directly by the \emph{is a} relation and guarantees that if a protein belongs to specific GO term, then it belongs also to each node in each path from that term to the root if \emph{is a} relations are followed. This property can be translated into consistency constraints as shown in Section \ref{sec:rules}.

\begin{figure}[H]
    \begin{subfigure}[]{1\textwidth}
    	\centering
		\includegraphics[scale=.4]{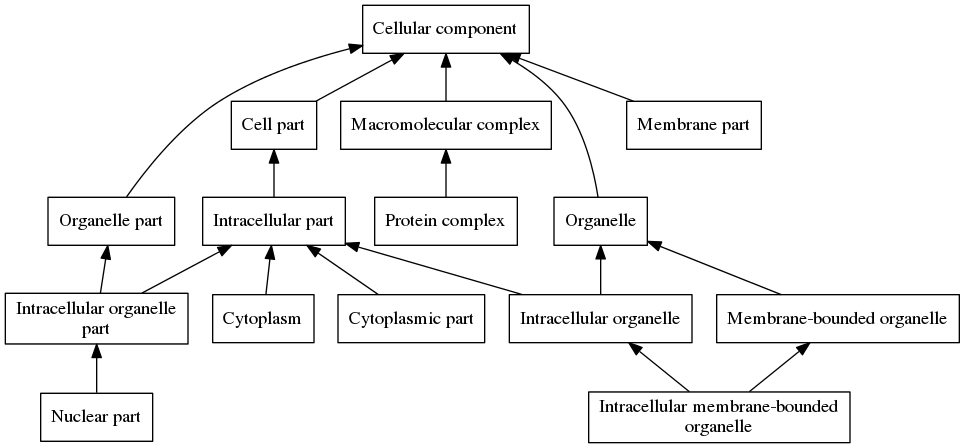}
        \caption{Cellular Component}
        \label{fig:CC}
	\end{subfigure}
    \vspace{10pt}
    \begin{subfigure}[]{1\textwidth}
    	\centering
		\includegraphics[scale=.4]{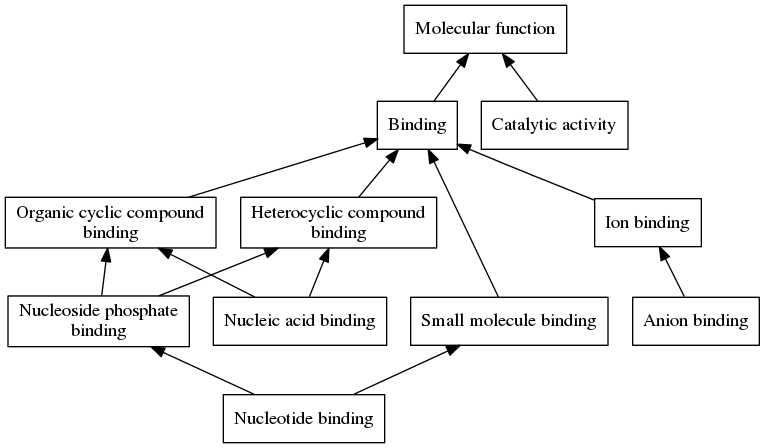}
        \caption{Molecular Function}
        \label{fig:MF}
	\end{subfigure}
\end{figure}

\begin{figure}
	\ContinuedFloat
    \begin{subfigure}[]{1\textwidth}
    	\centering
        \begin{turn}{90}
			\includegraphics[scale=.38]{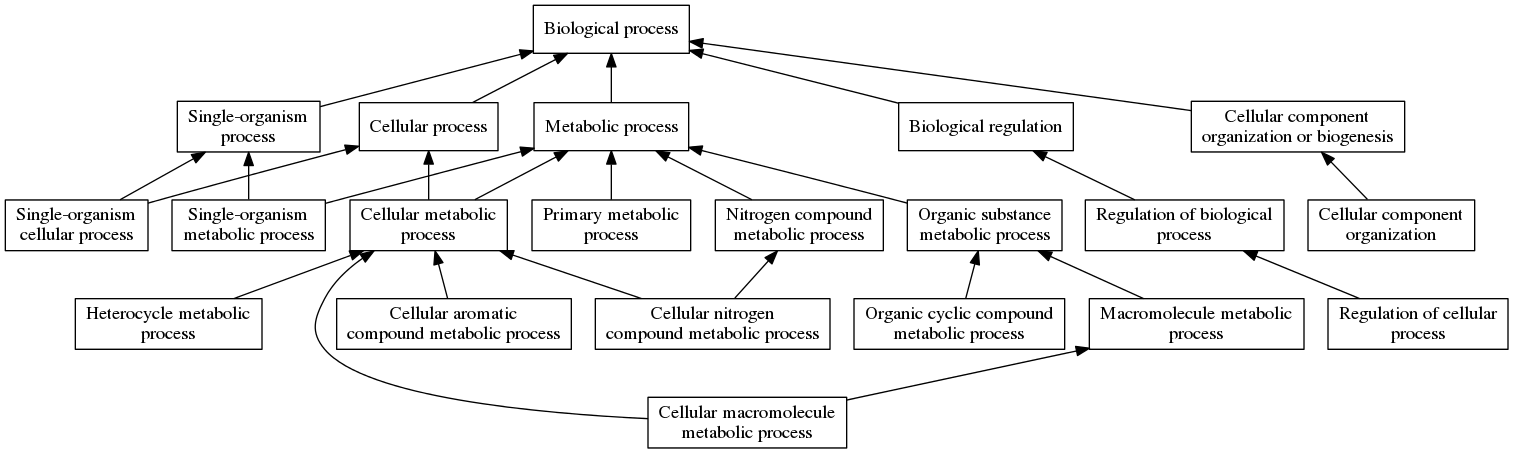}
		\end{turn}
        \caption{Biological Process}
        \label{fig:BP}
	\end{subfigure}

    \myCaption{Gene Ontology DAGs}{In this figure are reported three subgraphs of the first levels of Gene Ontology}
    \label{fig:awesome_image}
\end{figure}

\section{Bioinformatics approaches to the protein feature prediction problem}
\label{sec:related}
Literature concerning protein function prediction is really vast and the spectrum of methodologies even more. Attempts to identify the state of the art in this fields led to the CAFA\footnote{http://biofunctionprediction.org/} (Critical Assessment of Function Annotation experiment) challenge project, a worldwide effort aimed at analyzing and evaluating protein function prediction methods. A summary of the first competition can be found in \cite{CAFA_res} and points out that the abundance of biological data positively impacted on the results, showing big improvement with respect to traditional approaches like BLAST \cite{BLAST}, but there is no clear dominant methodology that overtakes the others. 

The goal of this section is to highlight the major classes of approaches used to face this problem. The methodologies taken into account are most relevant appeared in the CAFA competition whose details are extracted from an exhaustive survey done by \cite{pandey2006computational}. 

\subsection*{Sequence-based methods}

\subsubsection{Homology based}
In the lately 80's  the first sequence similarity systems like FASTA \cite{FASTA} and BLAST \cite{BLAST} make their appearance. Since then they have been widely applied to analyze nucleotidic and aminoacidic sequences from the major databases to find homologies in proteins.

Homology is a evolutionary concept that means some trait of two organisms derive from a common ancestor. Applying this concept to proteins we obtain that two proteins are homologous if they derive from a common ancestor protein.
The idea behind comparing amino acidic sequences is that highly conserved sequences may maintains their functionalities. Nonetheless two proteins deriving from a common ancestor (very similar sequence) may evolve different functions. This kind of proteins are called paralogous. It is the case for example of the Gal1 and Gal3 proteins of the {\it{S. cerevisiae}}. They have an identity 385 amino acids over 529 and a similarity of 443 amino acids over 529 but they cover completely different roles in the cell: Gal1 is a galactokinase (part of a metabolic pathway) while Gal3 is a transcriptional inducer \cite{platt2000insertion}.

\subsubsection{Motif based}
Motif-based methods are another sequence-based protein function annotation method. These techniques do not compare complete protein sequences. The investigated sequence is compared with known motifs, which are shorter signatures for protein domains, and for each of them is provided a statistic significance level. The commonly used measure is the Expected value or E-value and the P-value. The former represent the number of matches with the similarity  score obtained one can expect to find by chance in sequence of that size. The more it is close to zero, the more significant the match is. The latter instead measures the probability to find a matching with that similarity score by chance. The most popular database for retrieving specific information for motifs are probably Prosite \cite{PROSITE} and InterPro \cite{interpro}. Protein domains are well known to be strongly related to protein functions, especially at molecular level. Therefore a statistical significant match probably implies that the protein owns that particular function. 

Moreover motifs can be used to predict the cellular localization of a protein. During the translocation phase, special peptidic sequences are used to identify proteins and target them to their folding location. This information can be used as hint for their final sub cellular localization.

\subsection*{Structure based}
Protein structure is often more conserved then the sequence, but there are evidences that highly similar structures may show different functions. There are for example 27 homologous protein superfamilies folding in a {\it TIM barrel} shape that cover over 60 {\it EC numbers}\footnote{Enzyme Commission number is a numerical classification of enzymes, based on the reaction they catalyze.} \cite{greene2007cath}. There are some explanations to this aspect, the first is purely technical. Three dimensional protein structures are mainly obtain through X-ray crystallography, which requires, like the name says, a crystallized protein. Depending on the substrate in which the protein usually is folded, the crystallization may considerably alter the original shape of the protein. The second one is that, especially in catalytic activity, not only the shape of the active site is important but also the physico-chemical composition of the residues in the active site is crucial.

Nonetheless proteins with highly similar structures are likely to have the same or similar function. On this assumption rely many popular methods like {\it DALI} \cite{DALI}, {\it SSAP} \cite{SSAP}, {\it STRUCTURAL} and {\it CATHEDRAL} \cite{3D_struct}. 

\subsection*{Protein Interaction Networks}
Protein tends to perform their functions together with other proteins. It is often exactly this interaction that allows complex biological process. For example, myosin proteins by themselves would not be able to contract any muscle without the substrate made by the actin and this is just one of the many several examples of cooperation between proteins. This trend of proteins to co-work inspired some protein function prediction methods. Protein-Protein Interaction (PPI) networks are usually represented as graphs where proteins are the nodes and relationships the edges.

The underlying idea of this methods is that proteins that lie close in the PPI network are likely to share functions (especially at the biological process level). Indeed once the topology of the network is known, function annotations can be propagated to their neighbors or to the cluster they belong \cite{sharan2007network}. Unfortunately PPI networks are not easy to obtain and make these methods feasible just for well studies organism where the topology is at least partially known.

\subsection*{Gene expression data}
Gene expression experiments are a quantitative measure of the transcription rate of the mRNA during the protein synthesis. The most common technique to obtain information about genes expression level in a specific biological situation is the microarray analysis. Microarrays are silicon plates on which tens of thousands of probes are attached, where the probes are nucleotidic sequences corresponding to a specific gene\footnote{Probes have an high affinity to the target gene (complementary sequence of the mRNA), but in some cases also non-perfect matching could occur resulting in false positives.} which will interact with a marked cDNA\footnote{The mRNA sequences extracted from the cell are treated with the reverse transcriptase, which is an enzyme that generate a complementary DNA (cDNA) from an RNA template.}. Comparing data obtained by the analysis of different samples in various biological conditions, co-expression data can be inferred.

Assuming that co-expressed genes are likely to by biologically related and have part to the same biological process, the natural approach is to cluster them according to their expression profiles. Indeed, genes related to the same biological process are expected to be transcribes and silenced together. Function of the unannotated proteins are assigned according to the dominant function in which they are. This process can be summarized in the {\it guilt-by-association} concept.

\subsection*{Multiple Data Type}

Combining multiple data sources is a widely used technique in both, informatics and biology. The union of different information origins can improve the quality of the single dataset, indeed the errors of one can be compensated from another. Moreover data sources covering different aspects of the subject give a more comprehensive context to the analysis.

The conjunction of the different information can lead to a common format for all of them or to independent ones. An example of the former is the multiple kernel learning, where different kernel are used to build a final one in which sub-kernel are weighted according to their informativeness. Regarding the former, this work is a perfect example. Indeed the knowledge coming from the kernels is combined with the one of the constraints whose information origin is completely different.

\chapter{Methodological background}

\section{Kernel methods}
\label{sec:kernel}

Kernel methods are a class of machine learning algorithms for pattern analysis, whose best known exponents are the Support Vector Machines (SVM) \cite{SVM}. Thanks to their theoretical firmness, their computational efficiency and flexibility, kernel methods are ubiquitously presents, not only in the machine learning literature, but also in fields like computational biology where they are often used as black boxes to solve the most diverse tasks.

The common ground among these methods is the use of {\it kernel functions}. These functions allow to  operate in a high-dimensionality feature space, without the cost of computing the explicit inner product between feature vectors. This means that {\it kernel functions} merge the expressiveness of non linear machine learning algorithm with the efficiency of the linear ones. Intuitively {\it kernel functions} generalize the notion of dot product to arbitrary (even infinite-dimensional) input space and can be seen as a measure of similarity between objects. The flexibility and expressiveness of {\it kernel functions} arise from the fact that they can be reasonably easy applied not only to point in the euclidean space, but also to data structures, such as graphs, trees or sequences (extremely interesting in bioinformatics).

\begin{definition} {\bf (Valid kernel function \cite{generalized_representer})} Given a function $k:\XX \times \XX \rightarrow \real_{\geq 0}$ and a set of examples $\{x_1,\hspace{1pt}...\hspace{1pt},x_n\}$, the Gram matrix $K \in \real^{n\times n}$ is defined as:
\begin{equation}
K_{i,j}:=k(x_i,x_j)
\end{equation}
If the Gram matrix generated by $k$ is positive semi-definite, i.e. if it satisfies the condition:
\begin{equation}
\sum_{i,j} c_i c_j K_{i,j},\hspace{10pt}\forall\cc\in\real^n
\end{equation}
then $k$ is a valid kernel function.
\end{definition}

Valid {\it kernel functions} are very important in machine learning because they always correspond to a dot product in some Reproducing Kernel Hilbert Space (RKHS). Moreover the Representer Theorem \cite{generalized_representer} shows that problems in the form:
\begin{equation}
f^* = \underset{{f\in\mathcal{H}}}{\textrm{arg min}}\hspace{5pt}c((x_1,y_1,f(x_1)), ... , (x_n,y_n,f(x_n))~+~g(\lVert f \rVert)
\label{eq:representer_problem}
\end{equation}
where $\mathcal{H}$ is an appropriate RKHS, $g$ a strictly monotonically increasing real-valued function and $c$ an arbitrary cost function, admits representation in terms of kernel expansions of the form:
\begin{equation}
f^*(\xx)=\sum_{i=1}^n \alpha_i k(\xx, \xx_i)
\label{eq:representer_result}
\end{equation}
where $\alpha_i \in \real$ for all $i \in \{1,...,n\}$. The significance of this theorem is that it demonstrates that many learning methods allow solution that can be expressed as expansion in the training data.

{\bf Example} (Support Vector Machines\cite{SVM}) Given the labeled dataset $D=\{(\xx_1,y_1),\hspace{1pt}...\hspace{1pt},(\xx_n,y_n)\}$ with $y_i \in \{+1,-1\}$, the SVM classifier (without the bias\footnote{The biased version of the Representer theorem do not change the significance of the obtainable results, but adds technicalities} \cite{uSVM}) can be interpreted as a regularization method in which:
\begin{equation}
c((\xx_1,y_1,f(\xx_1)), ... , (\xx_n,y_n,f(\xx_n)) = \gamma \sum^n_{i=i} \textrm{max}\{0,1-y_i \ww^T \Phi(\xx_i)\},
\end{equation}
and $g(\lVert f \rVert) = \frac{1}{2} \lVert w \rVert^2$. $\Phi(\cdot)$ is the function mapping the input example in some feature space. By letting $\gamma \rightarrow \infty$ we obtain the Hard-margin version. Focusing on this last point we obtain:
\begin{equation}
\begin{split}
\underset{\boldsymbol{w}}{\textrm{min}}&\hspace{5pt} \frac{1}{2}\lVert \ww \rVert^2 \\
\textrm{subject to:}&\hspace{5pt}y_i \ww^T \Phi(\xx_i) \geq 1 \hspace{10pt} \textrm{for all }i \in {1,...,n}
\end{split}
\end{equation}
that is exactly the primal formulation of the hard-margin SVM. The constrain derives by infinitely penalizing misclassified samples.
Consider now the dual formulation of the problem:
\dualFormulation
where $\lambda_i$ for all $i \in \{1,...,n\}$ are the lagrangian multipliers. The instances for which $\lambda_i\alpha_i> 0$ are the {\it support vectors}. In this formulation the decision function will be in the form: $f(\xx)=\sum_{i=0}^n {\lambda_i y_i \Phi(\xx_i)^T \Phi(\xx)}$.
Data examples appear just in the form of $\Phi(\xx)^T\Phi(\xx')$ and can therefore be replaced by the {\it kernel function} $k(\xx,\xx')$. This technique is known as {\it "kernel trick"} and leads to this kernel formulation:
\dualKernelFormulation
In this case the decision function could be expressed as:
\begin{equation}
f(\xx)=\sum_{i=1}^n \alpha_i k(\xx, \xx_i)
\end{equation}
where $\alpha_i=\lambda_i y_i$. \newline\null\hfill $\blacksquare$

\subsection{Kernels in bioinformatics}

Computational biology and bioinformatics communities have extremely benefited from the introduction of kernel methods, developing on top of them a multitude of kernel functions from scratch or adapting already existing ones. In the following will be given some examples of kernel functions that has been successfully applied to bioinformatics tasks. 

\subsubsection{Spectrum and mismatch string kernels}
Spectrum string kernel functions, also known as $k$-mers kernels, are a family of kernel thought to compare sequences on the base of the number of their common contiguous sub-strings of length $k$ ($k$-mers). These methods have been proposed by Leslie et al. in \cite{spectrum, mismatch}.

Given a sequence $\mathcal{S}$ from an alphabet $\mathcal{A}$ of size $|\mathcal{A}|=l$ and a number $k\ge 1$, the $k$-spectrum of $\mathcal{S}$ is the set of all the subsequence of length k that it contains. Let $\Phi_k : \XX \rightarrow \real^{l^k}$ be the function mapping $\mathcal{S}$ in its feature space (all possible sequences of length $k$ from the alphabet $\mathcal{A}$). Then:
\begin{equation}
\Phi_k(\mathcal{S})=[\phi_\alpha(\mathcal{S})]_{\alpha \in \mathcal{A}}
\end{equation}
where $\phi_\alpha(\mathcal{S})$ is the number of times $\alpha$ occurs in $\mathcal{S}$. The k-spectrum kernel for sequences $\mathcal{S}_1$ and $\mathcal{S}_2$ is then
\begin{equation}
K_k(\mathcal{S}_1, \mathcal{S}_2) = \langle \Phi_k(\mathcal{S}_1),\Phi_k(\mathcal{S}_1) \rangle
\end{equation}
Depending on the size of the alphabet and the value of $k$ the vectors out coming from $\Phi_k(\mathcal{S})$ can be very sparse and therefore the dot product for computing the kernel very expensive. In practice is however not necessary to compute neither the explicit vectors nor the explicit dot product. Indeed, in \cite{spectrum} is proposed an efficient computation of the kernel based on suffix trees whose overall complexity is $O(n\textrm{log}(n))$ where $n$ is the length of the longest sequence.

In \cite{mismatch} Leslie et al. propose an alternative version of this kernel, the mismatch string kernel. This kernel expand the concept of string kernel by allowing up to $m$ mismatches when comparing two.

%Both these kernel are particularly effective in discovering remote homologies in protein sequences.

%\subsubsection{Tree Kernel}
%\cite{tree1,tree2}

%\subsubsection{Kernel on graphs}

%blA BLA	

%\subsubsection{Profile kernel}

%blA BLA	

\section{Semantic-Based Regularization}
\label{sec:sbrs}
Semantic-Based Regularization (SBR) is framework developed by \cite{SBRS} at the University of Siena, whose development stems from the need to incorporate prior knowledge into the well established machinery of statistical learning algorithms. The previous standard approach to introduce field-specific knowledge into the learning process was mainly focused on building sophisticated kernel functions, that could express the more information as possible. Unfortunately many interesting relationships between learned categories could not be easily incorporated into a kernel function. SBR approaches this problem allowing the user to inject prior information, in the form of First order Logic (FOL), that describes the relationships occurring between and inside the classes.

Semantic-based regularization, as the name may suggest, is formulated as multitask regularization problem on top of which is added a penalty cost coming from the infringement of the constraints (semantic part) as shown in Equation \ref{eq:sbrs}. The main feature of SBR are described in \cite{SBRS} and can be summarized in:
\begin{itemize}
  \item {\it Learning from constraints in kernel machines}\newline 
  The usual kernel methods machinery has been expanded in order to integrate the new concept of constraints, whose satisfaction degree is assessed through unsupervised data. This assessment allows SBR to refine his predictions trying to balance the label error of the training set and the constraints violation of the unsupervised examples.

  \item {\it Bridging logic and kernel machines}\newline T-norms allows the translation of first order logic clauses in real-valued functions that take part to the learning phase, ending up with a constrained multi-task optimization problem. The conversion of logic clauses into real values builds  a natural bridge between logic and kernel machines.

  \item {\it Stage-based learning}\newline Unfortunately, unlike classical kernel methods, constraints expressed in logic clauses increase the computational complexity, ending up in some cases with non-convex optimization problem. This makes typical optimization approaches not suitable. Therefore \cite{SBRS}, inspired by the work of Jean Piaget on stages of cognitive development, developed a stage-based gradient descent algorithm. This approach consists essentially in two phases. In the first one only supervised example are taken into account running the learning process until convergence without constraints, which are added only during the second phase. Being supervised example, at least in theory, coherent with the constraints, the first stage gives better starting point, rather than a random one, for the seconds stage gradient descending.
\end{itemize}

The ability to merge FOL constraints and multi-task kernel machines, makes SBR very attractive for bioinformatics  studies. It is the case of "Improved multi-level protein-protein interaction prediction with semantic-based regularization" of \cite{multi}. In this work constraints are used to improve the protein-protein interaction by imposing the need of two compatible domains (one per protein) and at least five interacting residues per protein. These boundaries refined the prediction increasing the consistency of the results. These are great hypothesis also for this work because they allow us to represent grounded biological information obtained from Gene Ontology as FOL rules achieving the consistency deriving from the True Path Rule. The actual constraints are explained in Methods.

\subsection{Theoretical foundation of SBR}
Let us consider a multitask learning problem, with $T$ the total number of tasks and where each $k$-th task is represented by a function $f_k$ defined in an appropriate Reproducing Kernel Hilbert Space (RKHS) $\mathcal{H}_k$.
In \cite{SBRS} is proposed a more general version of the Representer Theorem that extend it to multitask optimization problems. Indeed Equation \ref{eq:representer_problem} can be generalize to:
\begin{equation}
[f^*_1,...,f^*_T]= \underset{f_1 \in \mathcal{H}_1, ... , f_T \in \mathcal{H}_T}{\mathrm{arg min}} ~ E[f_1,...,f_T]
\end{equation}
and each function in the solution can be expressed in the form:
\begin{equation}
f^*_k(\xx_k)=\sum_{x^i_k\in S_k}\alpha_{k,i}^*K_k(\xx_k,\xx_k^i)
\end{equation}
where $K_k$ is the kernel corresponding the the space $\mathcal{H}_k$ and $S_k$ the set of available samples for the task $k$.

Given that the tasks have to satisfy a set of $H$ constraints the assumption made in SBR is that these task are correlated. The constrains are then defined by the functionals:
\begin{equation}
\phi_h:\mathcal{H}_1 \times ... \times \mathcal{H}_T \rightarrow [0,+\infty)
\end{equation}
and $\phi_h([f_1,...,f_T])=0$ for $h \in \{1,...,H\}$ (i.e. the constraints are satisfied) must hold for any valid choice of $f_k \in \mathcal{H}_k$ for $k \in \{1,...,T\}$. In Subsection \ref{sub:trans} we will show how the constraints are actually translated into real values.

The resulting optimization problem can be now expressed as a combination three component (Equation \ref{eq:sbrs}), the first penalizing the complexity of functions, the second penalizing the errors in the training set and the last one penalizing the infringement of the constraints.
\begin{equation}
\label{eq:sbrs}
\lambda_R \sum_{k=1}^T\lVert f_k \rVert ^2 + \sum_{k=1}^{T} \sum_{(x_k^i,y_k^i)\in \mathcal{L}_k} L(f_k(x_k^i),y^i_k)+\lambda_C \sum^H_{h=1} \phi_h(\mathcal{S},f)
\end{equation}
where $L$ is a loss function for the labeled examples and $\mathcal{S}$ is the set of examples available. The meta-parameters $\lambda_R$ and $\lambda_C$ impact respectively on the contribute of the regularization part and of the constraints one.
Representer theorem allows us to rewrite this optimization problem in term of kernel expansion. Let $f_k = \GG_k \aalpha_k$, then we obtain:
\begin{equation}
\lambda_R \sum_{k=1}^T \aalpha^T_k \GG_k \aalpha_k + \sum_{k=1}^{T} \boldsymbol{L}(\GG_k\aalpha_k,{\bf{y}}_k)+\lambda_C \sum^H_{h=1} \phi_h(\GG_1\aalpha_1,...,\GG_T\aalpha_T)
\end{equation}
where $\GG_k$ and $\aalpha_k$ are respectively the gram matrix and the weight vector for the task $k$.

The optimization is now in term of the weights vectors and can be done by descending the gradient. As described at the beginning of this section, SBR introduces a stage-based learning procedure by which the initial steps of the gradient descent are done with $\lambda_C=0$ in order to find a local minimum from where introducing the constraints. This heuristic is necessary because constraints in most interesting cases (such as this work) are non-linear and make therefore the problem non-convex.

\subsection{First Order Logic: an overview}
First Order Logic (FOL) is symbolical reasoning language in which statements are composed by logical concept (predicates) associated with logical subjects (variables) and quantifiers. It finds his main application field in the formal reasoning. FOL is a syntactical and semantical expansion of the proposition logic.

Propositional Logic (PL) is a relatively simple yet expressive formal language based on the concept of propositions, which given a truth assignment can be true or false. Propositions are logical formulas whose building blocks are proposition variables and logical connectives that are, in order of priority, $\lnot$ (not), $\land$ (and), $\lor$ (or), $\imply$ (imply) and $\bimply$ (double implication or equivalence). The table \ref{tab:truth} shows how the logical symbols modifies the values of the logical formulas $A$ and $B$.

\begin{table}[b]
\centering
\begin{tabular}{|c|c|c|c|c|c|c|}
\hline
$A$ & $B$ & $\lnot A$ & $A\land B$ & $A\lor B$ & $A\imply B$ & $A\bimply B$ \\ \hline
T   & T   & F         & T          & T         & T           & T            \\ \hline
T   & F   & F         & F          & T         & F           & F            \\ \hline
F   & T   & T         & F          & T         & F           & F            \\ \hline
F   & F   & T         & F          & F         & F           & T            \\ \hline
\end{tabular}
\myCaption{Logical connectives truth table}{}
\label{tab:truth}
\end{table}

Propositional logic performs extremely well for expressing specific concepts, but lacks of the possibility to bound a variable to a concept. For example the statement \emph{`Sara is happy, then she smiles'} can be easily expressed in PL ($sara\_is\_happy\imply sara\_smiles$), but statements like \emph{`There is a person named Sara'} can not be expressed in a compact and significative way in PL. First Order Logic extends propositional Logic with the aim to allow the association of a subject to a concept and their quantification.

Firs order logic inherits the logical symbols of PL to which it adds the quantifiers $\forall$ (for all) and $\exists$ (exists) and an infinite set of logical variables $x_1, x_2,...$. In addition to the logical symbols, FOL uses logical predicate $P(\cdot)$ which returns if the predicate is satisfied given the input. We are now able to express the statement \emph{`There is a person named Sara'} in FOL as $\exists x.Person(x) \land Named\_Sara(x)$ where $x$ is a logic variable, and $Person$ and $Named\_Sara$ are two logic predicate with arity one (number of parameters). If a grounding of the variable $x$ is given it is now possible to state whether the statement is satisfied.

First Order Logic is a powerful tool that will allow us to write complex biological relations in compact and unequivocal way.

\subsection{Translation of First Order Logic into real valued constraints}
\label{sub:trans}

The translation of a First Order Logic formula into a real value is fundamental for converting the constraints into a penalizing  term in the optimization problem. Without loss of generality the FOL formula can be rewritten moving all the quantifiers at its beginning (Prenex Normal Form).
\begin{equation}
\label{eq:example}
\underbrace{\forall x~ \forall y}_\text{\small \it Quantifiers} ~.~ \underbrace{(A(x) \land A(y)) \imply (B(x) \land B(y))}_\text{\small \it Quantifiers-free formula} 
\end{equation}
It can be noticed that given a grounding of the logic variables, the quantifiers-free part of the formula is a formula in propositional logic and can be therefore mapped in $[0,1]$ with one of the methodologies coming from the context of \emph{Fuzzy Logic}.

\subsubsection{Fuzzy Logic and T-norms}

\emph{Fuzzy Logic} is an extension of the Boolean logic in which the concept of true and false are `blurred'. Indeed to statements are assigned truth values ranging in $[0,1]$, where $0$ is completely false and $1$ completely true.

In order to operate in this logic the usual operands have to be reformulated to be able to handle real values. Notice that as in the boolean logic only two operators are necessary to obtain all the others. In this case we will use $\lnot$ and $\land$. Let $N:[0,1]\rightarrow[0,1]$ and $T:[0,1]\times[0,1]\rightarrow[0,1]$ respectively the mapping of $\lnot$ and $\land$. Then:
\begin{equation}
N(x_1) = 1 - x_1
\end{equation}
where $x_1$ is a fuzzy logic variable. It is clear that when $x_1$ approximates $0$ or $1$ the result is exactly what we would expect from the boolean logic.

The function $T$ generalizes the conjunction and correspond to a family of functions called \emph{T-norms} (triangular norms). \emph{T-norms} must satisfy the following properties: commutativity ( $T(x_1,x_2)=T(x_2,x_1)$), monotonicity ($T(x_1,x_2) \leq T(x_2,x_3)$ if $x_1 \leq x_2$ and $x_2 \leq x_3$), associativity: ($T(x_1, T(x_2, x_3)) = T(T(x_1, x_2), x_3)$) and the number 1 acts as identity element ($T(x_1, 1) = x_1$). All these properties which clearly agree with the boolean $\land$ ones.

Some important \emph{T-norm} examples (which are also implemented in SBR) are:
\begin{itemize}
\item Minimum $T$-norm: $T_{min}(x_1,x_2) = \textrm{min}\{x_1,x_2\}$
\item Product $T$-norm: $T_{prod}(x_1,x_2) = x_1\cdot x_2$
\item Lukasiewicz $T$-norm: $T_{Luk}(x_1,x_2) = \textrm{max}\{0, x_1 + x_2 -1\}$
\end{itemize}

For example the Product $T$-norm, used in most of the experiments will behave like this:
\begin{equation}
\begin{split}
\lnot x_1 ~~\overset{\small \it mapped}{\longrightarrow}& ~~N(x_1)=1-x_1 \\
x_1 \land x_2 ~~\overset{\small \it mapped}{\longrightarrow}& ~~T_{prod}(x_1,x_2)=x_1\cdot x_2 \\
x_1 \lor x_2 \equiv \lnot(\lnot x_1 \land \lnot x_2)~~\overset{\small \it mapped}{\longrightarrow}& ~~N(T_{prod}(N(x_1),N(x_2))) =\\
& ~~1-((1-x_1)(1-x_2))=\\
& ~~x_1+x_2 - x_1\cdot x_2\\
\end{split}
\end{equation}

Let us consider the logical implication $x_1 \imply x_2$. The natural approach would be to rewrite this formula as $\lnot x_1 \lor x_2$ and treat as seen above. This will end up in
\begin{equation}
\lnot x_1 \lor x_2 ~~\overset{\small \it mapped}{\longrightarrow} ~~ 1 + x_1 \cdot x_2 - x_1
\end{equation}
which corresponds to the implication but does not completely capture the inference process performed in
a probabilistic or fuzzy logic context. There exists indeed a binary function for each $T$-norm that plays the role of implication, the \emph{residuum}. Given two logical sub-formulas $x_1$ and $x_2$ in fuzzy logic the \emph{residuum} is defined as:
\begin{equation}
\begin{cases}1 & x_1 \leq x_2\\
R & x_1 > x_2
\end{cases}
\end{equation}
where $R$ is specific for each $T$-norm. In particular:
\begin{itemize}
\item Minimum $T$-norm Residuum: $R=x_2$
\item Product $T$-norm Residuum: $R=\frac{x_2}{x_1}$
\item Lukasiewicz $T$-norm Residuum: $R=1-x_1+x_2$
\end{itemize}
In the experimental section we used the Product $T$-norm Residuum when we need to translate an implication in constraints. 

It is interesting to notice that, being the constraints applied on unsupervised examples, once the variable are grounded, the values used for converting the constraints into real values are the ones coming from the decision function associated to the predicate\footnote{In SBR is also possible to mix learned information (values come from the decision functions) or given information. In this case the values used come from the labeling of the data}. Recovering the example used in Equation \ref{eq:example} the quantifiers-free part, using Product $T$-norm Residuum, can be mapped into:
\begin{equation}
\begin{dcases}
1 \vphantom{\frac{0}{0}} & f_A(x_1) \cdot f_A(x_2) \leq f_B(x_1) \cdot f_B(x_2)\\
\frac{f_A(x_1) \cdot f_A(x_2)}{f_B(x_1) \cdot f_B(x_2)} & otherwise
\end{dcases}
\end{equation}
where $f_A,f_B$ are the decision functions for the predicate $A$ and $B$.

\subsubsection{Quantifiers in SBR}
In addition to logical connectives results coming from \emph{fuzzy logic} and $T$-norms, SBR implements a mapping also for the quantifiers that are needed for expressing properties on the whole dataset.

The universal quantifier in SBR is a measure of the violation of the constraints on the dataset. Let $\forall v. E(v,\mathcal{P})$ be an universal quantified formula corresponding to the $h$-th constraints. Here $E$ is a logical expression  over the variable $v$ and the predicates $\mathcal{P}$, which, once $v$ is grounded, corresponds to a \emph{Fuzzy logic} formula and can therefore be mapped into a real value by the function $f_E(\xx,\mathcal{P})$, where $\xx$ represents the grounding of $v$ s.t. $\xx\in\mathcal{S}$. The mapping can be now expressed as:
\begin{equation}
\forall v.E(v,\mathcal{P}) ~~\overset{\small \it mapped}{\longrightarrow}~~ \phi_h(\boldsymbol{f}, \mathcal{S}) = \sum_{\xx \in \mathcal{S}} 1-t_E(\boldsymbol{f},\xx)
\end{equation}
In the general case where multiple universal quantifier can be expressed, summation are nested for each quantifier. In particular, the mapping for $\forall v_1...v_n.E(v_1...v_n,\mathcal{P})$ will be:
\begin{equation}
\phi(\boldsymbol{f}, \mathcal{S}) = \sum_{\xx_1 \in \mathcal{S}_1} ... \sum_{\xx_n \in \mathcal{S}_n} 1-t_E(\boldsymbol{f},\xx_1...\xx_n)
\end{equation}
Let us take for example the Equation \ref{eq:example}, the corresponding mapping will be:
\begin{equation}
\phi(\boldsymbol{f}, \mathcal{S}) = \sum_{\xx_1 \in \mathcal{S}_1} \sum_{\xx_2 \in \mathcal{S}_2}
\begin{dcases}
0 \vphantom{\frac{0}{0}} & {\small f_A(x_1) \cdot f_A(x_2) \leq f_B(x_1) \cdot f_B(x_2)}\\
1-\frac{\Large f_A(x_1) \cdot f_A(x_2)}{\large f_B(x_1) \cdot f_B(x_2)} & otherwise
\end{dcases}
\end{equation}

The existential quantifier express the presence of a variable that verifies the statement. In a \emph{Fuzzy logic} contest, the intuition is that the desired result would be to return the value of the most satisfied statement. Indeed, we can formulate $\exists v. E(v,\mathcal{P})$ as:
\begin{equation}
\phi(\boldsymbol{f}, \mathcal{S}) = \underset{\xx \in \mathcal{S}}{\textrm{min}} ~1 - t_E(\boldsymbol{f},\xx)
\end{equation}
that can be in a very natural way extended to the n-existential quantifier $\exists_n$ as:
\begin{equation}
\phi(\boldsymbol{f}, \mathcal{S}) = \sum_{\xx \in \mathcal{S}_{\textrm{ n min}}} 1 - t_E(\boldsymbol{f},\xx)
\end{equation}
where $\mathcal{S}_{\textrm{ n min}} = \underset{\xx \in \mathcal{S}}{\textrm{arg min}_n} ~ 1 - t_E(\boldsymbol{f},\xx)$, i.e. the first $n$ assignment of $\xx$ in the dataset that more satisfy the constraint. It is interesting to notice that taking the two extreme values for $n$ i.e. $n=1$ and $n=|S|$ we obtain respectively the existential quantifier and the universal one.

\chapter{Methods}

\section{Problem definition}

As mentioned in the introduction, protein function prediction is a central but yet hard task in bioinformatics. There are plenty of papers about this topic and each has a slightly different interpretation of the problem. We have now given the necessary tools for understanding the definition of our problem.  \vspace{-10pt}
\begin{definition}
\emph{\bf(Gene Ontology cut)} Consider the labeled dataset $D_\mathcal{L}=\{(p_1,S_1),$ $(p_2,S_2),...\}$ where $p_i$ is the $i$-th protein and $S_i$ the set of its annotation. Then, given a set of namespaces $\mathcal{N}$, a level threshold $l$ and a count threshold $c$, then we call $\hat{GO}$ a subgraph of GO such that:
\begin{equation}
\hat{GO}(D,\mathcal{N},l,c)=\{t\mid t \in \textrm{GO } \land level(t) \leq l \land |proteins(D,t)| \geq c\}
\end{equation}
where $level$ and $proteins$ are functions that return respectively the level and the proteins associated with a Gene Ontology term.
\end{definition}\vspace{-10pt}
This limitation is necessary for practical reasons. A minimum number of examples of each class in each fold is needed for a correct training and the level limit prevents to analyze too many predicates that would make the learning phase computationally too heavy.

The set of annotation $S_i$ must be consistent, that means that if the term $t\in \hat{GO}(\mathcal{N},l,c)$ belongs to $S_i$ then all its ancestors must be in $S_i$. This consistency property is inherited from the True Path Rule of Gene Ontology explained before.
Proteins can be described with a set of `fair' features, i.e. features that it is reasonable to believes that unseen proteins will be in posses of.

It is now possible to define our problem, both from a conceptual and from a more practical point of view.

\begin{definition}
\emph{\bf(Single Protein Function Prediction)} Given the labeled dataset $D_\mathcal{L}$, a GO cut $\hat{GO}(\mathcal{N},l,c)$ and a previously unseen protein $p$ with its features, the Single Protein Function Prediction (SPFP) problem consists in finding the consistent subgraph $\hat{S} \subseteq \hat{GO}(D,\mathcal{N},l,c)$ that most likely represents the set of the annotations of $p$.
\end{definition}

This is a typical definition of an inductive classification task, that clearly express the concepts aiming our work. However, SBR provides additional tools that we have exploited to improve the quality of the prediction. The problem can be therefore reformulated by adding the technical details as follows.

\begin{definition}
\emph{\bf(Multiple Protein Function Prediction)} Consider a labeled dataset $D_\mathcal{L}$ and a GO cut $\hat{GO}(\mathcal{N},l,c)$. Now, given a set of unseen proteins $D_\mathcal{U}$ and a set of constraints $\Phi$, the Multiple Protein Function Prediction (MPFP) problem consists in finding, for each $p\in D_\mathcal{U}$, the subgraph $\hat{S} \subseteq \hat{GO}(D,\mathcal{N},l,c)$ consistent with the constraints $\Phi$, that most likely represents the set of the annotations of $p$.
\end{definition}

The multiple protein set up allows us to use SBR in transductive learning mode. Moreover, as we have seen in Section \ref{sec:sbrs}, the unlabeled examples are fundamental to effectively impose constraints into the learning process.

\begin{figure}[ht]
\centering
\includegraphics[width=1\textwidth]{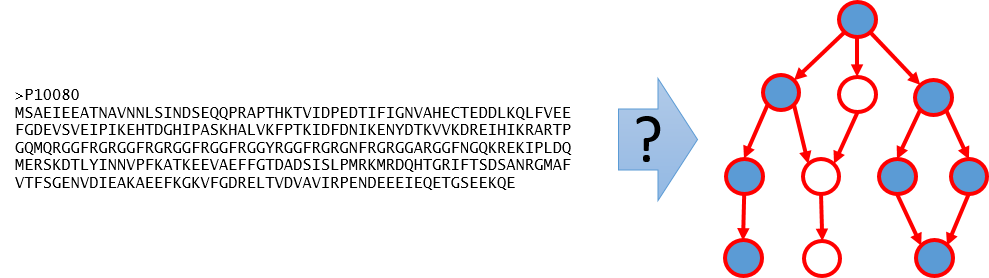}
\myCaption{General schema of the protein function prediction problem}{}
\end{figure}

\section{Kernels}
The choice of the kernel function is crucial for the outcomes of experiments conducted with Kernel Machines. The preliminary experiments made at the beginning of this project have been executed with the kernel of \cite{YIP}, which was designed for predicting protein-protein interactions. The obtained results made us clear the need of a set of kernel function specific for our task. Moreover the kernel used by \cite{YIP} contained as a feature the subcellular localization of the proteins, that could not be consider a completely fair feature for our task. Indeed the subcellular localization is an information hard to obtain in laboratory, therefore it is improbable that the unseen proteins, on which the method will be applied, will be in posses of this feature.

In this section we will propose the four kernel functions we used in for the experimental results.
In addition to the kernel used to measure the protein similarity, we used the kernel proposed in \cite{YIP} for measuring protein-pairs similarity in the experiments where also the Protein-Protein interaction has been exploited as a rule.
All the proposed kernel functions have been applied to our dataset, and for each of them we built the corresponding Gram matrix, which was used as input for the SBR experiments. In Appendix \ref{ap:kernel} are reported the heatmaps of the proposed kernels.

\subsubsection{Domains}
Domain-based kernel are known in the literature for well performing in Molecular Function prediction. Ours, in particular, is based on the data that can be retrieved from InterPro \cite{interpro} when an amino acid sequence is submitted.

InterPro offers a web service where amino acids sequence can be submitted and, via the \emph{HMMER3} \cite{HMMER3} algorithm, compared to the domain, family and superfamily data contained in its database. The results are exported in \emph{xml} format and for each protein are associated the corresponding information retrieved. Once all proteins $p_i$ have been analyzed, the Gram matrix for this kernel can be defined as:
\begin{equation}
G_{i,j} = \frac{|A_i \cap A_j|}{\sqrt[]{|A_i|^2\cdot|A_j|^2}}
\end{equation}
where $A_k$ is the set of annotation retrieved in InterPro for the $k$-th protein.

\subsubsection{Spectrum}
Spectrum kernel rely only on the amino acid sequence of the protein as described in Section \ref{sec:kernel}. In our experiment we used the normalized version of the string kernel function (no mismatch) shown in Section \ref{sec:kernel}:
\begin{equation}
k^*(p_i,p_j)= \frac{k(p_i,p_j)}{\sqrt[]{k(p_i,p_i)k(p_j,p_j)}}
\end{equation}
Being probably the simplest sequence-based we as a baseline for this kind of methods.

\subsubsection{Protein complexes}
The protein-complexes kernel is a diffusion kernel \cite{diffusion} %(see Section \ref{sec:kernel}) 
on the protein complexes graph of the yeast \cite{pu2009up}. As can be notice from Figure \ref{fig:heatmaps} (Appendix) this kernel is the most sparse, it is almost punctiform. This is due to the fact that protein complexes do not form a dense graph, it is indeed better described by the concept of graph forest, i.e. a set of (small) graphs not interconnected. The biological explanation of this behavior is that, despite not being fixed, protein complexes subunits are  usually part of just one complex and not exchanged between different complexes.

\subsubsection{Microarray}
The Microarray kernel matrix has been constructed computing the correlation for the gene expression data between each pair of proteins. Microarray data has been extracted from \cite{GE1,GE2} and covers the cell-cycle regulation and the response to environmental changes.

Let $X$ and $Y$ be the gene expression vectors for the protein $p_i$ and $p_j$ then the kernel function is:
\begin{equation}
k(p_i,p_j)= \frac{1}{n} \sum_{x\in X} \sum_{y\in Y} (x-\mu_{X})(y-\mu_{Y})
\end{equation}
where $n$ is the number of expression data per protein and $\mu_{X}$ the sample mean of $X$.

\section{SBR rules}
\label{sec:rules}
One of the fundamental contribute of this work is the integration of prior knowledge with the protein feature prediction process. Thanks to the Semantic-Based Regularization framework, we have been able to translate biological constraints coming from Gene Ontology and/or from Protein-Protein interaction network and incorporate them into the usual kernel methods machinery. In what follows are shown and explained the logical formulas that we used in our experiments.
\begin{figure}[b!]
\centering
\includegraphics[width=1\textwidth]{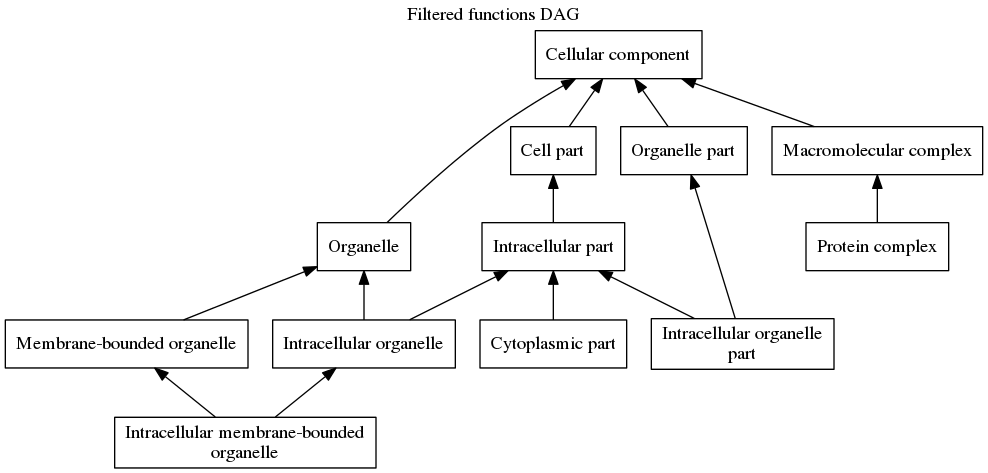}
\myCaption{Gene Ontology sub-graph}{}
\label{fig:cc_dag}
\end{figure}

\subsection{Ontology Consistency}
Gene Ontology is a trans-organism hierarchical ontology whose annotations are manually curated. These facts make GO a huge pool of information. Indeed we started exploiting the structure of the GO DAGs in order to force the consistency of the prediction, i.e. if an annotation $\alpha$ is predicted for a protein $p$, then $p$ should be predicted also in all the ancestors of $\alpha$. Let us take for example the sub-graph of the cellular component DAG of Gene Ontology shown in Figure \ref{fig:cc_dag}. If a protein $p$ is predicted to be part of an {\it intra-cellular organelle}, by extension $p$ is also part of all the GO terms bounded to the intra-cellular organelle one by an {\it is a} relation, that are {\it organelle} and {\it intra-cellular part}.
Moreover, assuming that a protein has at least one path from the root to one of the leaves, if a protein $p$ is predicted in a GO annotation $\alpha$ we can face two situations: 1) $\alpha$ is a leaf, base case, or 2) $\alpha$ is an inner node and that means that $p$ should be in at least one of the children of $\alpha$.
These two constraints are strictly related to the structure of Gene Ontology and try to guarantee that the consistency of the predictions, from which the name of {\it Ontology Consistency rules}.

Let $Prot$ be the set of proteins, $\mathcal{P}$ be the set of analyzed biological predicates (e.g. {\it catalytic activity, metabolic process, ...}), $chil: \mathcal{P} \rightarrow 2^\mathcal{P}$ a function that given a predicate return its children and $par:\mathcal{P}\rightarrow 2^\mathcal{P}$ a function that return the parents of a predicate. Given a predicate $u \in \mathcal{P}$, the first constrain cam be expressed as:
\begin{equation}
\forall x \in Prot ~\bigwedge_{p \in par(u)} u(x) \imply p(x)
\label{eq:fip}
\end{equation}
and the second as:
\begin{equation}
\forall x \in Prot~~u(x) \imply \bigvee_{c \in chil(u)} c(x)
\label{eq:pif}
\end{equation}
The translation into a SBR-readable logic formula is fairly straightforward. Formula \ref{eq:fip} is rewritten as a list of single constraints in the form of
$$\forall x \in Prot~\textrm{Q}(x) \imply \textrm{P}(x)$$
where P and Q are predicates s.t. $\textrm{P} \in par(\textrm{Q})$, for each parent of the predicate Q in order to avoid the conversion in the conjunctive normal form. Formula \ref{eq:pif} remain substantially identical.

\subsubsection{Bin nodes}
For practical reasons working on the whole GO DAGs is unfeasible. In fact a minimum number of protein per annotation is necessary for producing reasonable predictions. Therefore in our experiment the analyzed predicate have always been a subset of the GO annotations. This pruning however produces an undesirable effect that makes Formula \ref{eq:pif} no more valid. Indeed it is no more true that a protein associated with an inner-node of the GO DAG is at least associated to one of his children. It could be in fact the case that the pruned node was the only one associated with that protein.

In order to fix this drawback we introduce a class of nodes, whose goal is to gather all the proteins of the removed predicates, the {\it bin nodes}. Given a specific cut of the GO DAGs, the criteria for adding a {\it bin node} to annotation $\alpha$ are that $\alpha$ has at least one children node that survived to the cut and at least one who don't.

Let us expand the definitions of $par$ and $chil$ in such a way that $par:\mathcal{P}\rightarrow 2^{\mathcal{P} \cup \mathcal{B}}$ and $chil: \mathcal{P}\cup\mathcal{B} \rightarrow 2^\mathcal{P}$ where $\mathcal{B}$ is the set of {\it bin nodes}. With this relatively small hack we are able to prune the GO DAGs without loosing the consistency with the annotated dataset.

\subsubsection{Trans-hierarchy relations}
Since the first experiments, it was clear that some kernel performed way better on some GO DAG rather then an other. It is, for example, the case of the domain-based kernel whose performance on the molecular function DAG are considerably better when compared with the one obtained on the biological process DAG. We investigate therefore a way to propagate the information between the hierarchies. Gene Ontology structure came in our help providing a small but promising set of relations that could occur even between the GO DAGs. In particular these are \emph{occurs in}, \emph{part of} and \emph{regulates}.
\begin{itemize}
\item The \emph{occurs in} relation is a relationship that binds the biological process and the cellular component DAGs and therefore is not really useful in our experiment. Moreover there are just few relationships of this kind in the Gene Ontology.
\item The \emph{regulates} relation is relatively abundant and occurs between the hierarchies analyzed in this work. Unfortunately biological regulation takes place in a multitude ways that makes the translation of this relation into logical formula almost impossible.
\item The \emph{pat of} relation instead is a good candidate for the trans hierarchy-rules, i.e. it occurs between molecular function and biological process DAGs with the meaning that a molecular function is part of some biological process.
\end{itemize}
The \emph{part of} relation has therefore translated into the implication
$$
\forall x \in Prot \textrm{ Q}(x) \imply \textrm{P}(x)
$$
for all predicate P, Q such that $\textrm{P}\in part\_of(\textrm{Q})$.

\subsection{Protein-Protein Interaction}
In order expand the knowledge base we tried to integrate the Protein-Protein Interaction (PPI) data into the learning process. This information is partially already present into the complex kernel, but for the other kernels we decided to inject it in form of SBR constraints.

The motivations behind the idea of using this information can be found in the PPI network methods (Section \ref{sec:related}). The underlying concept is that interacting proteins are likely to share their functions. From this idea we implemented the \emph{PPI rules}. So, let BOUNDP be the binary predicate related to the interaction of proteins then:
\begin{equation}
\label{eq:false}
\forall x,y \in Prot \textrm{ BOUND}(x,y) \imply \textnormal{P}(x) \bimply \textnormal{P}(y)
\end{equation}
for all P $\in \mathcal{P}$. The translation in natural language of this constraint is that if two proteins interact then they share the predicate, for all the predicates.

Unfortunately this constraints is from a biological point of view not completely true. In fact it is true that interacting protein tends to share functions, but is usually not true that they share all of them. This results was also confirmed by the statistics gathered on the dataset. Table \ref{tab:pp1} and Table \ref{tab:pp2} show that, excluding the roots of the GO DAGs, there is no predicate that fully verifies the constraint \ref{eq:false}. Moreover in Table \ref{tab:jac} can be observed
that Biological Process is more related to the interaction when compared to Molecular Function.

\begin{table}[tp]
\centering
\begin{tabular}{@{}lccc@{}}
\toprule
                   & Average & Median & $\sigma$ \\ \midrule
Biological Process & 0.629   & 0.615  & 0.274  \\
Molecular Function & 0.560   & 0.500  & 0.317  \\ \bottomrule
\end{tabular}
\myCaption{Jaccard coefficient on pairs of interacting proteins}{The table shows the average, the median and the standard deviation of the Jaccard coefficient computed for each interacting pair of proteins in the dataset on the annotation set restricted to the analyzed GO DAG with cut 2,100}
\label{tab:jac}
\end{table}

These results can be appreciated also from a biological point of view. Indeed, two interacting proteins very likely share a biological process, but it is often not true that they are accomplishing the same molecular function. For example, muscle contraction is triggered by the arrival of a neurotransmitter in the neuromuscular junction, that connects the nervous systems and the muscular system via synapses. In the neuromuscolar synapse there is the Nicotinic acetylcholine receptor, which is a protein complex composed by five subunits, arranged symmetrically around a central pore. Two of them, called $\alpha$-subunits, are the only one able to bind the neurotransmitter (small molecule binding activity). The other three subunits have just structural roles. This is an example of proteins interacting in the same biological process, which does not share any molecular function.

According to the results obtained from the statistics gathered for the previous constraints we decided to weaken the constraints and reduce it to the following constraint:
\begin{equation}
\forall x,y \in Prot \textrm{ BOUND}(x,y) \imply \bigvee_{P\in\restr{\mathcal{P}}{BP}} \textnormal{P}(x) \cap \textnormal{P}(y)
\end{equation}
where $\restr{\mathcal{P}}{BP}$ are the predicate examined with that are in the Biological Process DAG. We call this constraint DPP constraint.

Both the PP and the DPP constraint have been applied in combination with the OC rules and using given BOUND (true values), which are called PP1 and DPP1, or by learning them, PP2 and DPP2.

\section{Measuring performances}
\label{sec:perf}

For a robust assessment of the performance of our setup we used a 10-fold inner cross-validation procedure. The pseudo code of the fold generation algorithm is reported in Algorithm \ref{alg:folds}. In order to be consistent with the original task the final statistics are computed only over the investigated predicates, i.e. the \emph{bin nodes} have been excluded being technical artifacts. The performance of the prediction of protein interaction are also taken as separated statistics.

Measuring performance in a multi-labeled setting is not trivial. There is indeed the new concept of \emph{partially correct} that is not present in binary or multi-class problems \cite{multi_metrics1}. The results can analyzed from two different point of view, and from each of these we different implementations of the common metrics  used for assessing performances.

Some notation:
\begin{itemize}
\item $n$: the number of examples in the test set (we are using a cross validation and therefore $n=|D|$)
\item $k$: the number of possible labels (the number of predicates analyzed in the experiment)
\item $Y_i$: the set of label of the $i$-th example (in our case the set of annotations for the $i$-th protein)
\item $Z_i$: the set of predicted label for the $i$-th example
\item $TP_j,~TN_j,~FP_j,~FN_j$: respectively the true positives, true negatives, false positives and false negatives of the $j$-th predicate.
\end{itemize}

\subsubsection{Example based}
Example based evaluation measures analyze goodness of the results with respect to the examples. The extreme example of these class of metrics is the \emph{Exact Match Ratio} which return the ratio of completely correctly classified example with respect to the whole test set. Of course this measure too aggressive, especially for a hard task like this one.

We evaluated our results according to the metrics proposed by \cite{eval2} that are:
$$\textnormal{examples precision }P^\epsilon = \frac{1}{n} \sum_{i=1}^k \frac{|Y_i \cap Z_i|}{|Z_i|}$$
The \emph{example precision} $P^\epsilon$ is the proportion of correctly predicted labels to the total of predicted  labels, averaged over all examples.
$$\textnormal{examples recall }R^\epsilon = \frac{1}{n} \sum_{i=1}^k \frac{|Y_i \cap Z_i|}{|Y_i|}$$
The \emph{example recall} $R^\epsilon$ is the proportion of correctly predicted labels to the total of correct labels, averaged over all examples.
$$\textnormal{examples }F1^\epsilon = \frac{1}{n} \sum_{i=1}^k \frac{2|Y_i \cap Z_i|}{|Y_i|+|Z_i|}$$
The \emph{example F1} $F1^\epsilon$ is the harmonic mean of $P^\epsilon$ and $R^\epsilon$.

An important statistics for this work is the measure of the consistency of the prediction for each example. It is indeed important to be able to quantify the improvements dictated by the introduction of constraints. Unfortunately the task is very specific to our problem and there are no used metrics, so we defined ourselves the consistency metrics as:
\begin{equation}
C^\epsilon=\frac{1}{n}\sum_{p\in Prot}~\frac{1}{|\mathcal{P}(p)|}\sum_{n \in \mathcal{P}(p)}
\begin{dcases*}
1 \vphantom{\frac{0}{0}} & level(n) = 1\\
\frac{|parents(n)\cap\mathcal{P}(p)|}{|parents(n)|} & \textrm{otherwise}
\end{dcases*}
\end{equation}
where $parents$ and $level$ are respectively the functions returning the set of parents and the level in GO of a given predicate and $\mathcal{P}$ the function returning the predicted predicates for a protein. This metric is completely independent from the true values of the annotations, it just measures how much the consistency constraints are verified. The measure is unfortunately influenced by the level threshold used for the GO cut. It is indeed harder for deeper DAGs rather then for the ones with few levels to obtain high values for this metric. In the experimental phase it has been indeed used just to measure the changes inside a single experiment caused by rules.

\subsubsection{Label based}
Label based metrics focus themselves on the goodness of the predicate predictions. In this case we can apply the typical binary metrics and then compute the average between the predicates. There are two possible strategies for computing the average: 1) the \emph{macro} average that simply computes the mean between the metrics for each predicate or 2) the \emph{micro} average. In this case the confusion matrices for each predicate are summed up and on the obtained matrix are calculated the statistics.
Here the definitions of the metrics used:
$$
P^\lambda_\mu = \frac{\sum_{i=1}^{k}TP_i}{\sum_{i=1}^{k} TP_i+FP_i} \hspace{40pt}
P^\lambda_M= \frac{1}{k}\sum_{i=1}^{k}\frac{TP_i}{TP_i+FP_i}
$$
$$
R^\lambda_\mu = \frac{\sum_{i=1}^{k}TP_i}{\sum_{i=1}^{k}TP_i+FN_i}
 \hspace{40pt}
R^\lambda_M = \frac{1}{k}\sum_{i=1}^{k}\frac{TP_i}{TP_i+FN_i}
$$
$$ 
F1^\lambda_\mu = \frac{2\sum_{i=1}^{k}TP_i}{\sum_{i=1}^{k}2TP_i+FP_i+FN_i} \hspace{40pt}
F1^\lambda_M = \frac{1}{k}\sum_{i=1}^{k} \frac{2TP_i}{2TP_i+FP_i+FN_i}
$$
There is no trivial choice of the best averaging method, so we have computed both.

In addition to the numerical metrics, we developed an algorithm for plotting macro averaged precision-recall curves. This kind of plot is very useful to understand the trend of the results, regardless to the decision threshold. Unfortunately there is no direct definition of the precision recall curve for  multi-label problems, but \cite{multi_metrics1} suggest that the definition of the usual metrics used for binary problems can be extended to the multi-label case by averaging the results. We decide to use a \emph{macro} average for this task because it comes more natural for computing the mean between curves.

The core idea of the algorithm we developed is to find, for each sampling coordinate on the recall axis, the best possible coordinate on the precision axis for each curve and then averaging them. So, let $PRC_p$ the precision-recall curve for a predicate $p\in\mathcal{P}$ expressed a couple of list of points $PRC_p = ([P_1,...,P_n],[R_1,...,R_n])$, where $(P_i,R_i)$ are respectively the the precision and recall coordinates of the $i$-th point int the list. Now suppose to have a set of $PRC$ curves $\mathcal{C}$ and $n$ sampling points, then the average curve can be computed with the following algorithm:

\begin{algorithm}[H]
\caption{\emph{Macro} averaged Precision-Recall Curve algorithm}
\label{alg:curves}
\FUNCTION{\textnormal{\emph{Macro} averaged Precision-Recall Curve (set $\mathcal{C}$, int $n$)}}{
	$p \gets [0]_n$ \tcp*{$(n+1)$-dimensional array filled with zeros}
    $r \gets [i/n \mid i \in \{0,...,n\}]$ \tcp*{$(n+1)$-dimensional array where the $i$-th position is $i/n$}
    \ForEach{$c_p,c_r \in\mathcal{C}$}{
        \For{$k \in \{0,...,n\}$}{
          $i \gets \textnormal{the index of the left nearest point in } c_p \textrm{ to }r[k]$\;
          $j \gets \textnormal{the index of the right nearest point in } c_p\textrm{ to }r[k]$\;
          $f \gets \textnormal{function of the line passing through } (c_p[i],c_r[i])\textnormal{ and }(c_p[j],c_r[j])$\;
          $p[k] \gets p[h] + f(r[k]) / |\mathcal{C}|$ \tcp*{divides the contribute by the size of $\mathcal{C}$}
    	}
    }
    \RETURN{p,r}
}
\end{algorithm}

In Figure \ref{fig:curves} (Appendix) is shown how the algorithm performs when an increasing number of curves is averaged. It is interesting to notice how well the algorithm approximate the first curve, meaning that our approach is performing reasonably good. Despite the variability of the curves, the macro averaged one can be still considered representative of the experiment.

\section{Workflow description}
\label{sec:work}

During the development of the project, we implemented a pipeline for automatizing the execution of new experiments. The code has been written in python with the use of the following libraries: pygraphviz \cite{pygraphviz} (plotting of the statistic and result trees) and scikit-learn \cite{scikit} (computing AUC metrics). In the following are described the most important steps of the developed pipeline.

\subsubsection{Dataset selection and annotation}

The dataset used in this work is a subset of the one used in \cite{multi}, which consists in is 1681 proteins of the \emph{S. cerevisiae}. In order to be used in our experiments, the protein must have at least one non-electronically-inferred annotation in each of the analyzed GO domains that are Molecular Function and Biological Process. Unfortunately some of the 1681 initial proteins were not suitable for being used, therefore the dataset is dynamically adapted to fit the constraints. Despite this, the dataset is per-level consistent, i.e. experiments executed on the same depth level  of the GO hierarchy use the same dataset.

\subsubsection{Gene Ontology data structure}
The obo-formatted files obtained from the Gene Ontology website has been initially parsed with {\it goatools}\footnote{https://github.com/tanghaibao/goatools}, an open-source python library created by Haibao Tang. During the implementation of the project this library has been widely modified to fit our needs.

Firstly, {\it goatools} takes into account only the \emph{is a} relations between the GO term, but in our experiments we wanted to exploit different relations in order to expand the knowledge base. Therefore the data structure has been modified for keeping track of all the possible relations. After that, parametric accessors methods has been added to allow simple filtering of the related nodes.
All the experiments of this work have been executed on a subset of the GO DAG hence a parametric pruning function has been added. This function takes as parameter the level and the count threshold and the GO namespaces to analyze. It returns a subset of the original GO DAG and for each GO term is associated the corresponding list of protein of the data set.

All these improvements proposal will be submitted to Haibao Tang at the end of the project.

\subsubsection{Fold generation for cross validation}

\begin{algorithm}[tb]
\caption{Folds generation algorithm}
\label{alg:folds}
\FUNCTION{\textnormal{generate folds (int $n$, set $\mathcal{P}roteins$, list $GO~\mathcal{T}erms$, function $f:\mathcal{P}\rightarrow 2^{\mathcal{T}}$, function $g:\mathcal{T}\rightarrow 2^{\mathcal{P}}$)}}{
	$\mathcal{T} \gets$ sorted $\mathcal{T}$ according to $|f(t)|$ for $t \in \mathcal{T}$\;
    list {\it folds} $\gets [\{\}]_n$\;
    set {\it selected} $\gets \{ \}$\;
    \ForEach{$t ~\in~\mathcal{T}$}{
    	{\it to select} $\gets$ $g(t) \smallsetminus ${\it selected}\;
        \ForEach{$p\in${\it to select}}{
        	int {\it index} $\gets \underset{i}{\textrm{argmin}}\{|folds[i]|\mid i \in 1,...,n\}$\;
            ${\it folds[index]} \gets {\it folds[index]} \cup \{p\}$\;
            $selected \gets selected \cup \{p\}$\;
        }
    }
    \RETURN{\it folds}
}
\end{algorithm}
The SBR software does not provide an internal cross validation assessment, so it has to be done externally by splitting the data set in $n$ folds and submitting to the software each time $n-1$ folds as training set and one as test set.

Our experiments are intrinsically multi-label (each protein can be associated to more annotations), therefore producing $n$ folds in where each class is balanced is very challenging. Trying to find a global optimum for all the possible GO term -our classes- led to an intractable problem. So we decided to generate the folds experiment by experiment with a greedy approach that exploits the structure of the dataset. This led to a flexible and light system whose results well fits our goals.

The main idea behind the algorithm is to have a sufficient number of positive example of each class in each fold. The most critical classes are obviously the ones with less positive entries, which, in our setting, correspond to the deepest nodes in the GO hierarchy. Once the leaf examples are completely splitted into the folds, for the True Path Rule (discussed previously), the parent nodes are consequently splitted. 

For these reasons the algorithm sorts the classes according to their positive example number in ascending order and starts splitting the first class. Then it continues iteratively until all the classes are separated in folds. The pseudocode in Algorithm \ref{alg:folds} shows the details of the algorithm.%and the performance summary is reported in the Results section.

\subsubsection{Execution}

The training and testing of each fold is executed by the SBRS framework. The execution time grows fast with number of the rules, hence multiple SBRS executions are launched in parallel to soften this bottleneck. Each of the $n$ execution produces a results file containing all the prediction values of the validation set that will be process afterwards.

\subsubsection{Postprocessing}

\begin{figure}[htb]
\centering
\includegraphics[width=1\textwidth]{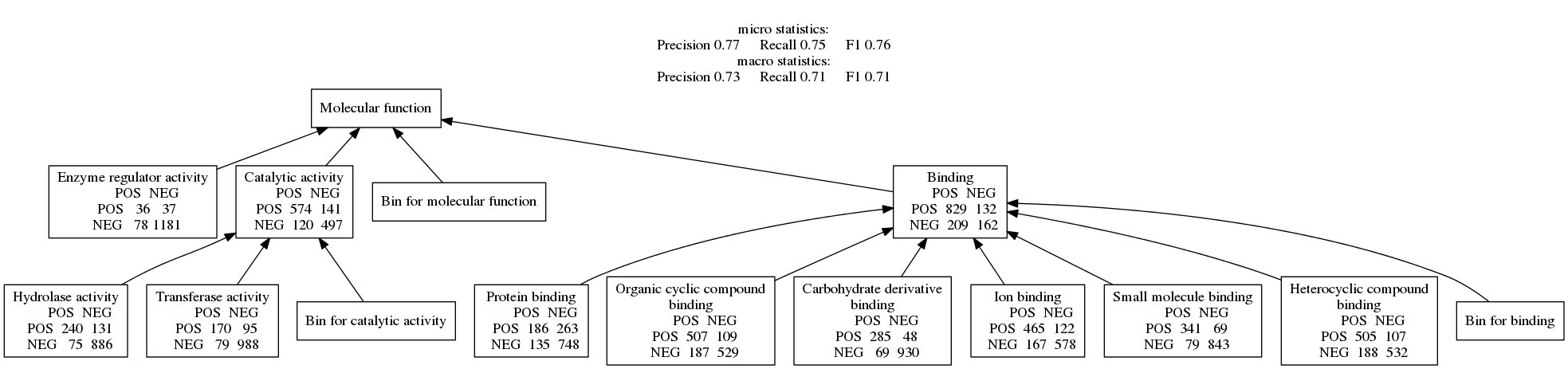}
\myCaption{Result tree example}{The Central Dogma of Molecular Biology describes the three fundamental processes for life. Is interesting to notice that all these are carried out by proteins.}
\label{fig:result_tree}
\end{figure}

The final results are gathered and post processed according to the methodologies showed in methods. The statistics are stored in text format and plotted on the corresponding GO DAG for highlighting the dependencies between the nodes. An example is shown in Figure \ref{fig:result_tree}. At last experiments parameters and result metrics are stored in a local database.

\chapter{Results}

The experimental phase of this work has been conducted on a subset (see Section \ref{sec:work} for the details) of the dataset used by \cite{YIP}, which consists of more then 1500 proteins of the \emph{Saccharomyces cerevisiae} (baker yeast). The \emph{S. cerevisiae} is a unicellular organism that belongs to the kingdom of \emph{Fungi}. The baker yeast is a model organism for \emph{eukaryotes}, it has the complexity of \emph{eukaryotes} but grows easily in laboratories and has a short life cycle. Due to this reasons, \emph{S. cerevisiae} was the first \emph{eukaryotic} organism to be completely sequenced (in 1996 by \cite{yeast}). Since then it has been widely studied and many specific resources can be find on the internet. Baker yeast represents therefore a perfect candidate for experimenting our setting.

As explained in Section \ref{sec:perf}, we performed a $10$-fold cross validation for the assessment of the performance for the different rule sets and kernels in the analyzed Gene Ontology cuts. Unfortunately the computational time for each experiment is considerable high (in some cases it takes more then two hours for the training of each fold). Therefore we were not able to perform an exhaustive meta-parameter search to optimize the performances. We decided so to keep the default meta-parameter of SBR for each experiment, i.e. $\lambda_R=1$ and $\lambda_C=1$ (see Section \ref{sec:sbrs}. 

The presentation of the results and the relative discussion will be divided according to the two analyzed DAGs in Gene Ontology, Biological Process and Molecular Function. Only at the end we will show the results obtained combining them in a joint prediction task. The division comes from the fact that this hierarchies have deep differences from a biological point of view and therefore the tested kernels behave highly differently in the two circumstances. In each section, the discussion will be organized according to the following pattern.
\newpage

\begin{itemize}
\item {\bf Baseline results} \newline
This portion of the discussion highlights the behavior of the kernel on the analyzed predicates in absence of rules. In this configuration SBR reduces to a $l_2$ regularized SVM task, indeed in absence of constraints, the learning problem becomes convex and converges to the global optimum.
This analysis is fundamental for a complete understanding of the results obtained with the introduction of rules.

\item {\bf Ontology Consistency constraints} \newline
The injection of Ontology Consistency (OC) constraints aims to improve the consistency of the predictions, with the final goal to refine the results through the propagation of information.

Consider the simple OC rule $h := \forall x. P_2(x) \imply P_1(x)$ where $P_2$ is a child node of $P_2$ in GO. Recall the definition of product residuum t-norm shown in Section \ref{sec:sbrs}, then the penalty function $\phi$ for $h$ is:
\begin{equation}
\phi(h,\mathcal{S}) = 1-\sum_{x \in \mathcal{S}}
\begin{dcases}
1 & P_2(x) \leq P_1(x)\\
\frac{P_1(x)}{P_2(x)} & otherwise
\end{dcases}
\end{equation}
where $\mathcal{S}$ is the set of unsupervised example on which the predictions for the predicates are computed. Consider now the case in which we have a strong positive prediction for the predicate $P_2$ and a slightly negative one for $P_1$. We know, from the True Path Rule of Gene ontology, that if a term is predicted, then all its ancestors should be predicted too. Indeed the constraints $h$ will contribute to the refinement of the results until $P_2(x) \leq P_1(x)$ will be satisfied. From this scenario the possible results on $P_1$ are:
\begin{itemize}
\item The positive prediction of $P_2(x)$ is correct (True Positive) and in the constrained gradient descent process the prediction of $P_1(x)$ is refined until the reaching of a value very close to $P_2(x)$ (low contribute) or greater of $P_2(x)$ (no contribute). In both cases $P_1(x)$ changed its prediction to positive, hence a True Positive.
\item The positive prediction of $P_2(x)$ is incorrect (False Positive) and in the constrained gradient descent process the prediction of $P_1(x)$ is forced to a positive classification value. In this case the action of the constraint badly impact on the performances. We end up indeed with two False positive, instead of just one. Nonetheless we are achieving a better consistency in the results.
\end{itemize}

The goal of this portion of the discussion is indeed to analyze the effects of this propagation of information on the predictions. As just explained, there are no guarantees that a better consistency will be reflected in a better quality of results.

\item {\bf Protein-Protein Interaction constraints} \newline
The goal of these sets of rules is to increase the knowledge base injecting information on the PPI into the learning process. As explained in the Section \ref{sec:rules} we applied these rules only on biological process to achieve biological consistency. In particular we used two supersets of constrains which differ by the base assumptions of the rules. In the first one indeed, the interacting proteins are supposed to share all biological processes. On the other hand, the second one assumes that two interacting proteins should be involved in at least one common biological process. The second one is more valid from a biological point of view, but less powerful. We will propose also the protein-protein interaction results to understand whether or not this predictions are influenced by the kernel used for the protein function prediction.

\end{itemize}

\subsection*{Notation summary}
We will propose here a short summary of the notation used in this chapter to facilitate the reading and the comprehension of the results.

\vspace{-10pt}
\paragraph{Rules} (See Section \ref{sec:rules})\vspace{-12.5pt}
\begin{itemize}
\item $\mathcal{E}$ : no rules
\item OC : Ontology Consistency rules
\item PPI : Protein Protein Interaction rules
\item DPPI : DNF Protein Protein Interaction rules
\end{itemize}

\vspace{-15pt}
\paragraph{Metrics} (See Section \ref{sec:perf})\vspace{-12.5pt}
\begin{itemize}
\item $Y^\epsilon,~Y^\lambda$ : respectively experiment- and label-based metrics
\item $Y_\mu,~Y_M$ : respectively micro- and macro-averaged metrics
\item $P,~R,~F1$ : respectively precision, recall and F1 measure
\item $\mathcal{C}$ : consistency measure
\end{itemize}

\vspace{-15pt}
\paragraph{Gene Ontology} (See Section \ref{sec:GO})\vspace{-12.5pt}
\begin{itemize}
\item BP : Biological Process hierarchy
\item MF : Molecular Function hierarchy
\item $H~_{l,c}$ : hierarchy cut with level threshold $l$ and count threshold $c$
\end{itemize}

\newpage

\section{Biological Process}
In this section we will propose and discuss the results obtained with our system using as label space the Biological Process DAG of Gene Ontology. In particular we limited the set of analyzed annotation to the GO cut with count threshold $c=100$ and level threshold $l\in\{2,3\}$. We will focus on the impact of the different kernels and sets of rules in this situation.

The list of the analyzed predicate and the complete result tables can be found in the Appendix.

\subsection*{Baseline results}
\begin{table}[H]
\centering
\begin{tabulary}{\textwidth}{clCCCCCC}
\toprule
Level               & Kernel                          & $P^\lambda_\mu$                        & $R^\lambda_\mu$               & $F1^\lambda_\mu$                       & $P^\lambda_M$                          & $R^\lambda_M$                 & $F1^\lambda_M$                         \\ \midrule
                    & Complexes                       & 0.460                                  & \textbf{0.932}                & 0.616                                  & 0.435                                  & \textbf{0.920}                 & 0.557                                  \\
                    & \cellcolor[HTML]{EFEFEF}Domains & \cellcolor[HTML]{EFEFEF}\textbf{0.575} & \cellcolor[HTML]{EFEFEF}0.682 & \cellcolor[HTML]{EFEFEF}\textbf{0.624} & \cellcolor[HTML]{EFEFEF}\textbf{0.520}  & \cellcolor[HTML]{EFEFEF}0.656 & \cellcolor[HTML]{EFEFEF}\textbf{0.558} \\
\multirow{-2}{*}{2} & Microarray                      & 0.453                                  & 0.568                         & 0.504                                  & 0.440                                  & 0.576                         & 0.460                                   \\
                    & Spectrum $k=3$                  & 0.569                                  & 0.536                         & 0.552                                  & 0.496                                  & 0.458                         & 0.457                                  \\ \midrule
                    & Complexes                       & 0.352                                  & \textbf{0.928}                & 0.510                                  & 0.332                                  & \textbf{0.914}                & 0.458                                  \\
                    & \cellcolor[HTML]{EFEFEF}Domains & \cellcolor[HTML]{EFEFEF}0.493          & \cellcolor[HTML]{EFEFEF}0.705 & \cellcolor[HTML]{EFEFEF}\textbf{0.58}  & \cellcolor[HTML]{EFEFEF}\textbf{0.442} & \cellcolor[HTML]{EFEFEF}0.681 & \cellcolor[HTML]{EFEFEF}\textbf{0.517} \\
\multirow{-2}{*}{3} & Microarray                      & 0.361                                  & 0.575                         & 0.443                                  & 0.341                                  & 0.565                         & 0.396                                  \\
                    & Spectrum $k=3$                  & \textbf{0.498}                         & 0.487                         & 0.492                                  & 0.421                                  & 0.399                         & 0.396                                  \\ \bottomrule
\end{tabulary}
\myCaption{Baseline results for Biological Process}{The table shows the result metrics for the experiments without rules in the Biological Process DAG.}
\label{tab:bp_base}
\end{table}

We can appreciate from Table \ref{tab:bp_base} that the complex-based kernel and the domains one outperform the others in absence of constraints. Despite having similar $F1$ measures (especially within level 2) their behavior is very different. Indeed the complex kernel has an extremely high recall (more then 90\%), but a very modest precision. Domains kernel instead behaves more uniformly.

\begin{figure}[!b]
\centering
	\begin{subfigure}[]{0.48\textwidth}
    	\centering
		\includegraphics[width=.9\textwidth]{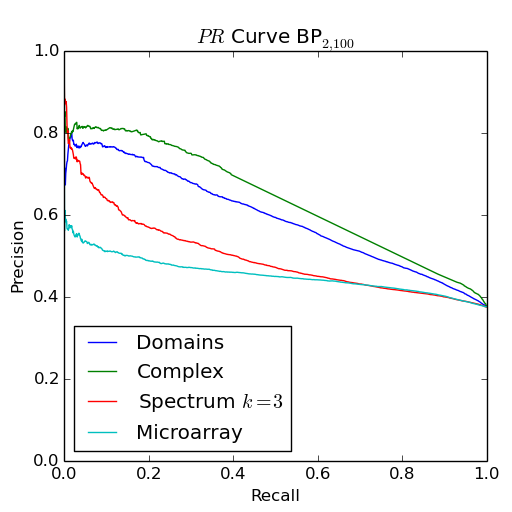}
	\end{subfigure}
    \begin{subfigure}[]{0.48\textwidth}
    	\centering
		\includegraphics[width=.9\textwidth]{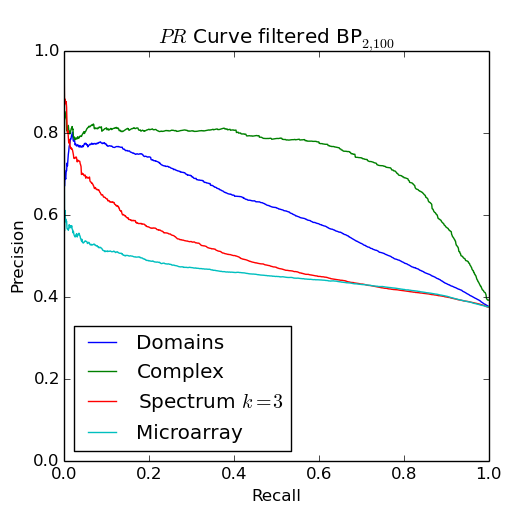}
	\end{subfigure}
\myCaption{Precision-Recall Curve of the kernel functions for Biological Process $_{2,100}$}{In the two plots are reported the PR curves obtained on BP$_{2,100}$ with the empty set of rules. The right plot highlights the effect of the filtering on the curves}
\label{fig:comp_BP}
\end{figure}

The unbalancing in the complex kernel is explained by the fact that the information that it contains is very sparse (see heatmap in Appendix Figure \ref{fig:complex}). From the learning process point of view this is translated into the impossibility to take any decision for proteins whose kernel matrix row is zero in each entry, biologically translated, the proteins for which there are no interaction information. The effect on the results is that many protein predictions converge in the prior, which is also the threshold. As convention the threshold is considered positive and therefore these examples are considered positive predictions, hence the high recall and low precision.
Differently, if the `undecided' proteins are left apart, we end up with the results shown in Table \ref{tab:bp_base_u}. Here the situation changes drastically and the complex kernel shows its potential. Indeed, with a $F1_\mu$ around the $80\%$ widely overtakes the the rivals. Figure \ref{fig:comp_BP} highlights this behavior. The Precision-Recall curve of the complexes kernel remains stable over the $70\%$ of precision until reaching almost a recall of $80\%$.

\begin{table}[t]
\centering
\begin{tabulary}{\textwidth}{clCCCCCC}
\toprule
Level & Kernel & $P^\lambda_\mu$ & $R^\lambda_\mu$ & $F1^\lambda_\mu$ & $P^\lambda_M$ & $R^\lambda_M$ & $F1^\lambda_M$ \\ \midrule
 & \cellcolor[HTML]{EFEFEF}Complexes & \cellcolor[HTML]{EFEFEF}\textbf{0.795} & \cellcolor[HTML]{EFEFEF}\textbf{0.839} & \cellcolor[HTML]{EFEFEF}\textbf{0.816} & \cellcolor[HTML]{EFEFEF}\textbf{0.722} & \cellcolor[HTML]{EFEFEF}\textbf{0.804} & \cellcolor[HTML]{EFEFEF}\textbf{0.754} \\
\multirow{-2}{*}{2} & Domains & 0.637 & 0.645 & 0.641 & 0.572 & 0.616 & 0.577 \\
% & Microarray & 0.453 & 0.568 & 0.504 & 0.44 & 0.576 & 0.46 \\
% & Spectrum $k=3$ & 0.569 & 0.536 & 0.552 & 0.496 & 0.458 & 0.457 \\
\midrule
 & \cellcolor[HTML]{EFEFEF}Complexes & \cellcolor[HTML]{EFEFEF}\textbf{0.734} & \cellcolor[HTML]{EFEFEF}\textbf{0.833} & \cellcolor[HTML]{EFEFEF}\textbf{0.781} & \cellcolor[HTML]{EFEFEF}\textbf{0.656} & \cellcolor[HTML]{EFEFEF}\textbf{0.79} & \cellcolor[HTML]{EFEFEF}\textbf{0.709} \\
\multirow{-2}{*}{3} & Domains & 0.572 & 0.671 & 0.617 & 0.516 & 0.647 & 0.562 \\
%& Microarray & 0.361 & 0.575 & 0.443 & 0.341 & 0.565 & 0.396 \\
% & Spectrum $k=3$ & 0.498 & 0.487 & 0.492 & 0.421 & 0.399 & 0.396 \\ 
\bottomrule
\end{tabulary}
\myCaption{Filtered baseline results for Biological Process}{This table shows the filtered version of the previously proposed ones. The undecided entries have been removed from the computation of the statistics, highlighting therefore the informative results}
\label{tab:bp_base_u}
\end{table}

Unfortunately we can not take this table as absolute value, because depending on how undecided values are treated we will definitively loose in precision or in recall, but it confirms the hypothesis that protein complexes are strongly related to biological processes. On the other hand it is interesting to notice that protein domains are, as one could expect, just marginally related to biological process. It is indeed true that proteins with similar domains may be part of completely different and not overlapping biological processes.

Moreover, even though it has been shown that gene expression data are correlated to biological processes \cite{GE3,GE4,GE5,GE6}, the microarray kernel performs poorly, even worse than the spectrum kernel. This may be partially explained by the combination of the noise that the corresponding heatmap highlights and the suboptimal parameter used in the experiments. A better choice of the $\lambda_R$ meta-parameter could indeed partially soften the effect of the noise in the data.

\begin{table}[t]
\centering
\begin{tabulary}{\textwidth}{clCCC|CCC}
\toprule
Level & Kernel & $P^\epsilon$ & $R^\epsilon$ & $F1^\epsilon$ & $f~P^\epsilon$ & $f~R^\epsilon$ & $f~F1^\epsilon$ \\ \midrule
 & \cellcolor[HTML]{EFEFEF}Complexes & \cellcolor[HTML]{EFEFEF}0.547 & \cellcolor[HTML]{EFEFEF}\textbf{0.941} & \cellcolor[HTML]{EFEFEF}\textbf{0.692} & \cellcolor[HTML]{EFEFEF}\textbf{0.812} & \cellcolor[HTML]{EFEFEF}\textbf{0.855} & \cellcolor[HTML]{EFEFEF}\textbf{0.833} \\
 & Domains & \textbf{0.604} & 0.688 & 0.643 & 0.631 & 0.652 & 0.641 \\
\multirow{-2}{*}{2} & Microarray & 0.469 & 0.578 & 0.518 & 0.469 & 0.578 & 0.518 \\
 & Spectrum $k=3$ & 0.573 & 0.549 & 0.560 & 0.573 & 0.549 & 0.560 \\ \midrule
 & \cellcolor[HTML]{EFEFEF}Complexes & \cellcolor[HTML]{EFEFEF}0.467 & \cellcolor[HTML]{EFEFEF}\textbf{0.940} & \cellcolor[HTML]{EFEFEF}\textbf{0.625} & \cellcolor[HTML]{EFEFEF}\textbf{0.759} & \cellcolor[HTML]{EFEFEF}\textbf{0.855} & \cellcolor[HTML]{EFEFEF}\textbf{0.804} \\
 & Domains & \textbf{0.535} & 0.707 & 0.609 & 0.566 & 0.673 & 0.615 \\
\multirow{-2}{*}{3} & Microarray & 0.367 & 0.585 & 0.451 & 0.367 & 0.585 & 0.451 \\
 & Spectrum $k=3$ & 0.503 & 0.500 & 0.501 & 0.503 & 0.500 & 0.501 \\ \bottomrule
\end{tabulary}
\myCaption{Example-based metrics on baseline results}{The prefixed $f$ stays for filtered.}
\label{tab:bp_base_exp}
\end{table}

Table \ref{tab:bp_base_exp} reports the example-based metrics for these experiments, with and without the filtering of the undecided examples. The example-based metrics differ from the label-based one by the fact that the former measure the quality of the prediction per protein, the latter per predicate. The overall results (no filtering) highlight the differences in the behavior of the two typologies of metric. Being the average of the results obtained per single protein, they do not take into account the differences of the annotation set sizes.

\subsection*{Ontology Consistency constraints}

\begin{table}[h]
\centering
\begin{tabulary}{\textwidth}{lCCC}
\toprule
               & $\mathcal{C}_\mathcal{E}$ & $\mathcal{C}_{\textrm{OC}}$ & \% increment \\ \midrule
Complexes      & 0.2006                  & 0.2006              & 0.000        \\
Domains        & 0.2482                  & 0.2537              & 2.198        \\
\cellcolor[HTML]{EFEFEF}Microarray     & \cellcolor[HTML]{EFEFEF}\textbf{0.3204}& \cellcolor[HTML]{EFEFEF}\textbf{0.4359} & \cellcolor[HTML]{EFEFEF}\textbf{36.03}\\
Spectrum $k=3$ & 0.0815                  & 0.1083              & 32.77        \\ \bottomrule
\end{tabulary}
\myCaption{Consistency of the predictions with OC rules on BP $_{2,100}$}{The table shows the comparison of the consistency measure $\mathcal{C}$ (see Section \ref{sec:perf}) in the experiments with and without consistency rules. The last column is the percentage increase obtained with the rules.}
\label{tab:BP_con}
\end{table}

Table \ref{tab:BP_con} proposes the values relative to the Ontology Constraints infringement metric $\mathcal{C}$. We can easily notice the difference in the impact of the rules on the kernel. Recall that in the experimental phase no meta-parameter tuning has been done and these results has been obtained therefore with the default meta-parameter of SBR. In particular, in this paragraph the $\lambda_C$ parameter plays a crucial role giving more or less importance to the constraints, and a suboptimal choice deeply impacts on the results. We can indeed notice the different behaviors of the kernels. Complexes and domains kernel are just minimally affected by the constraints. In these kernel in fact, the information is very concentrated and precise. In both cases, the constraints meta-parameter is too soft to induce a substantial variation. On the other hand, in microarray and spectrum kernel information is much more diluted and the $\lambda_C$ is sufficient to highly influence the results. As explained before, the fact that the constraints are more respected do not imply that the results will benefit in terms of prediction performance. The hope is that the dominant part of the implication is correct and therefore also the information propagated.
\begin{table}[t]
\centering
\begin{tabulary}{\textwidth}{clCCCCCC}
\toprule
Rules                        & Kernel                                 & $P^\lambda_\mu$                        & $R^\lambda_\mu$                        & $F1^\lambda_\mu$                       & $P^\lambda_M$                          & $R^\lambda_M$                 & $F1^\lambda_M$                         \\ \midrule
                             & Microarray                             & 0.453                                  & \textbf{0.568}                         & 0.504                                  & 0.440                                  & \textbf{0.576}                & 0.453                                  \\
\multirow{-2}{*}{$\mathcal{E}$} & \cellcolor[HTML]{EFEFEF}Spectrum $k=3$ & \cellcolor[HTML]{EFEFEF}\textbf{0.569} & \cellcolor[HTML]{EFEFEF}0.536          & \cellcolor[HTML]{EFEFEF}\textbf{0.552} & \cellcolor[HTML]{EFEFEF}\textbf{0.496} & \cellcolor[HTML]{EFEFEF}0.458 & \cellcolor[HTML]{EFEFEF}\textbf{0.569} \\ \midrule
                             & Microarray                             & 0.475                                  & \textbf{0.621}                         & 0.538                                  & 0.439                                  & \textbf{0.594}                & 0.475                                  \\
\multirow{-2}{*}{OC}         & \cellcolor[HTML]{EFEFEF}Spectrum $k=3$ & \cellcolor[HTML]{EFEFEF}\textbf{0.601} & \cellcolor[HTML]{EFEFEF}\textbf{0.621} & \cellcolor[HTML]{EFEFEF}\textbf{0.611} & \cellcolor[HTML]{EFEFEF}\textbf{0.496} & \cellcolor[HTML]{EFEFEF}0.491 & \cellcolor[HTML]{EFEFEF}\textbf{0.601} \\ \bottomrule
\end{tabulary}
\myCaption{Effect of the constraints on Biological Process $_{2,100}$}{}
\label{tab:BP_oc}
\end{table}

Table \ref{tab:BP_oc} shows an extract of the comparison between the results obtained with and without OC rules. The complexes and domains kernels have been excluded, because unchanged from a practical point of view. The complete results can be found in Appendix. It is easy to notice that the OC constraints not only impacted positively on the consistency of the prediction, but also on the quality. We get a $12.5 \%$ improvement in the $F1^\lambda_\mu$ for the spectrum kernel and a $6.5 \%$ for the microarray one.

\subsection*{Protein-Protein Interaction constraints}
\begin{table}[H]
\centering
\begin{tabulary}{\textwidth}{CCCCCCC}
\toprule
Rules & $P^\lambda_\mu$ & $R^\lambda_\mu$ & $F1^\lambda_\mu$ & $P^\lambda_M$ & $R^\lambda_M$ & $F1^\lambda_M$ \\ \midrule
$\mathcal{E}$ & 0.575 & 0.682 & 0.624 & 0.520 & \textbf{0.656} & 0.558 \\
OC & 0.575 & 0.682 & 0.624 & 0.521 & \textbf{0.656} & 0.559 \\
PPI & 0.573 & 0.679 & 0.621 & 0.519 & 0.655 & 0.556 \\
\rowcolor[HTML]{EFEFEF} 
DPPI & \textbf{0.625} & \textbf{0.698} & \textbf{0.660} & \textbf{0.544} & 0.637 & \textbf{0.577} \\ \bottomrule
\end{tabulary}
\myCaption{Effect of the OC rules on Biological Process $_{2,100}$}{}
\label{tab:BP_pp}
\end{table}
Table \ref{tab:BP_pp} reports an extract of the effects of the protein-protein interaction rules. We focus our attention on the domain kernel because it is the one best performing, but analogous consideration can be applied to all the others. Moreover the complete results can be found in the Appendix. It can be noticed that the set of rules that obtained the highest $F1$ scores is the DPPI one. Unfortunately, these results has been obtained with a different version of SBR, which admits DNF constrains formulas and uses different t-norms, and therefore non completely comparable with the previous examples. Nonetheless we can appreciate the improvement compared to the PPI rules, which worsen the overall prediction. As reported in Table \ref{tab:jac}, the PPI constraints are often violated already in the training set. The scarce validity of the rule leads to a flow of misleading information that inevitably will worse the prediction quality.

The prediction of protein-protein interaction has not been influenced by the underlying protein kernel. Indeed, we obtained the same results with all of them, meaning that we have been able to achieve a flow of information from the interactions to biological process but not vice versa. For the interaction prediction we obtained a $F1$ measure of circa $76\%$, both with the PPI constraints and with the DPPI.

\section{Molecular Function}

This section will propose the results for the Molecular Function DAG,  we will focus especially on the GO cut for MF with count threshold $c=100$ and level threshold $l \in \{2,3,4\}$. In these experiments we considered also the fourth level of gene ontology, because the MF DAG with the count threshold we used is smaller and it was therefore computationally easier to analyze deeper levels. As for Biological Process, the analysis will be centered around the effects of different kernel functions and sets of rules. 

\subsection*{Baseline results}

In Table \ref{tab:MF_empty} we can clearly notice how the domain-based kernel widely outperforms the competitors. It shows a good balance in precision and recall and achieves a $F1_\mu$ measure of almost $75\%$. The obtained results confirm the hypothesis that protein domains are strongly related to molecular function.
It is interesting to notice that the spectrum kernel, despite its simplicity, with a $65\%$ of $F1_\mu$ at level 2 has reasonably good performances. The explanation of this behavior can be found in the consideration made for the domains kernel. It is indeed true that the spectrum kernel is capable to detect conserved regions of sequences and therefore mimic the behavior of the domain kernel. From this observation we can conclude that molecular functions are easier to predict from protein sequences rather than biological processes.
\begin{table}[t]
\centering
\begin{tabulary}{\textwidth}{clCCCCCC}
\toprule
Level               & Kernel                          & $P^\lambda_\mu$                        & $R^\lambda_\mu$               & $F1^\lambda_\mu$                       & $P^\lambda_M$                          & $R^\lambda_M$                 & $F1^\lambda_M$                         \\ \midrule
                    & Complexes                       & 0.415                                  & \textbf{0.884}                & 0.565                                  & 0.404                                  & \textbf{0.869 }               & 0.527                                  \\
                    & \cellcolor[HTML]{EFEFEF}Domains & \cellcolor[HTML]{EFEFEF}\textbf{0.700} & \cellcolor[HTML]{EFEFEF}0.805 & \cellcolor[HTML]{EFEFEF}\textbf{0.749} & \cellcolor[HTML]{EFEFEF}\textbf{0.650} & \cellcolor[HTML]{EFEFEF}0.783 & \cellcolor[HTML]{EFEFEF}\textbf{0.701} \\
\multirow{-2}{*}{2} & Microarray                      & 0.437                                  & 0.565                         & 0.493                                  & 0.418                                  & 0.545                         & 0.452                                  \\
                    & Spectrum $k=3$                  & 0.549                                  & 0.798                         & 0.650                                  & 0.491                                  & 0.677                         & 0.555                                  \\ \midrule
                    & Complexes                       & 0.347                                  & \textbf{0.874}                & 0.496                                  & 0.337                                  & \textbf{0.861}                & 0.464                                  \\
                    & \cellcolor[HTML]{EFEFEF}Domains & \cellcolor[HTML]{EFEFEF}\textbf{0.665} & \cellcolor[HTML]{EFEFEF}0.829 & \cellcolor[HTML]{EFEFEF}\textbf{0.738} & \cellcolor[HTML]{EFEFEF}\textbf{0.625} & \cellcolor[HTML]{EFEFEF}0.827 & \cellcolor[HTML]{EFEFEF}\textbf{0.704} \\
\multirow{-2}{*}{3} & Microarray                      & 0.363                                  & 0.571                         & 0.444                                  & 0.353                                  & 0.561                         & 0.412                                  \\
                    & Spectrum $k=3$                  & 0.520                                  & 0.730                         & 0.608                                  & 0.477                                  & 0.624                         & 0.525                                  \\ \midrule
                    & Complexes                       & 0.305                                  & \textbf{0.869}                & 0.452                                  & 0.297                                  & \textbf{0.861}                & 0.423                                  \\
                    & \cellcolor[HTML]{EFEFEF}Domains & \cellcolor[HTML]{EFEFEF}\textbf{0.641} & \cellcolor[HTML]{EFEFEF}0.841 & \cellcolor[HTML]{EFEFEF}\textbf{0.727} & \cellcolor[HTML]{EFEFEF}\textbf{0.604} & \cellcolor[HTML]{EFEFEF}0.839 & \cellcolor[HTML]{EFEFEF}\textbf{0.695} \\
\multirow{-2}{*}{4} & Microarray                      & 0.324                                  & 0.562                         & 0.411                                  & 0.314                                  & 0.560                         & 0.384                                  \\
                    & Spectrum $k=3$                  & 0.496                                  & 0.696                         & 0.579                                  & 0.463                                  & 0.595                         & 0.504                                  \\ \bottomrule
\end{tabulary}
\myCaption{Baseline results for Molecular Function}{}
\label{tab:MF_empty}
\end{table}

Complexes- and microarray-based kernel do not seem to carry much information, but we know from the biological process analysis that the complex kernel suffers from a convergence of results in the threshold which deeply alter the statistics. We applied also here the filtering on the results. The results are reported in Table \ref{tab:MF_empty_u}.

\begin{table}[b]
\centering
\begin{tabulary}{\textwidth}{clCCCCCC}
\toprule
Level & Kernel & $P^\lambda_\mu$ & $R^\lambda_\mu$ & $F1^\lambda_\mu$ & $P^\lambda_M$ & $R^\lambda_M$ & $F1^\lambda_M$ \\ \midrule
\multirow{2}{*}{2} & Complexes & 0.622 & 0.676 & 0.648 & 0.568 & 0.635 & 0.590 \\
 & \cellcolor[HTML]{EFEFEF}Domains & \cellcolor[HTML]{EFEFEF}\textbf{0.829} & \cellcolor[HTML]{EFEFEF}\textbf{0.790} & \cellcolor[HTML]{EFEFEF}\textbf{0.809} & \cellcolor[HTML]{EFEFEF}\textbf{0.782} & \cellcolor[HTML]{EFEFEF}\textbf{0.765} & \cellcolor[HTML]{EFEFEF}\textbf{0.770} \\\midrule
% & Microarray & 0.437 & 0.565 & 0.493 & 0.418 & 0.545 & 0.452 \\
% & Spectrum $k=3$ & 0.549 & 0.798 & 0.650 & 0.491 & 0.677 & 0.555 \\ 
\multirow{2}{*}{3} & Complexes & 0.566 & 0.650 & 0.605 & 0.510 & 0.606 & 0.548 \\
 & \cellcolor[HTML]{EFEFEF}Domains & \cellcolor[HTML]{EFEFEF}\textbf{0.824} & \cellcolor[HTML]{EFEFEF}\textbf{0.817} & \cellcolor[HTML]{EFEFEF}\textbf{0.821} & \cellcolor[HTML]{EFEFEF}\textbf{0.799} & \cellcolor[HTML]{EFEFEF}\textbf{0.813} & \cellcolor[HTML]{EFEFEF}\textbf{0.803} \\ \midrule
% & Microarray & 0.363 & 0.571 & 0.444 & 0.353 & 0.561 & 0.412 \\
% & Spectrum $k=3$ & 0.520 & 0.730 & 0.608 & 0.477 & 0.624 & 0.525 \\
\multirow{2}{*}{4} & Complexes & 0.531 & 0.639 & 0.580 & 0.482 & 0.600 & 0.530 \\
 & \cellcolor[HTML]{EFEFEF}Domains & \cellcolor[HTML]{EFEFEF}\textbf{0.823} & \cellcolor[HTML]{EFEFEF}\textbf{0.831} & \cellcolor[HTML]{EFEFEF}\textbf{0.827} & \cellcolor[HTML]{EFEFEF}\textbf{0.803} & \cellcolor[HTML]{EFEFEF}\textbf{0.827} & \cellcolor[HTML]{EFEFEF}\textbf{0.813} \\\bottomrule
% & Microarray & 0.324 & 0.562 & 0.411 & 0.314 & 0.560 & 0.384 \\
% & Spectrum $k=3$ & 0.496 & 0.696 & 0.579 & 0.463 & 0.595 & 0.504 \\ 
\end{tabulary}
\myCaption{Filtered baseline results for Molecular Function}{This table reports the filtered version of the results presented in the previous one.}
\label{tab:MF_empty_u}
\end{table}
Here we can further appreciate the performances of the domains-based kernel. Indeed, removing from the analysis the entries converged to the threshold, despite loosing a little recall, the precision is greatly increased reaching more than $82\%$ for all the analyzed thresholds. This is translated into an improvement of the $F1_\mu$ measure, which not only is stable over the $80\%$, but it increase with number of level analyzed. This particular is very interesting because it implies that the prediction of deeper nodes in the molecular function hierarchy are more related to domains than the higher ones.
\begin{figure}[t]
\vspace{-10pt}
\centering
	\begin{subfigure}[]{0.45\textwidth}
    	\centering
		\includegraphics[width=.9\textwidth]{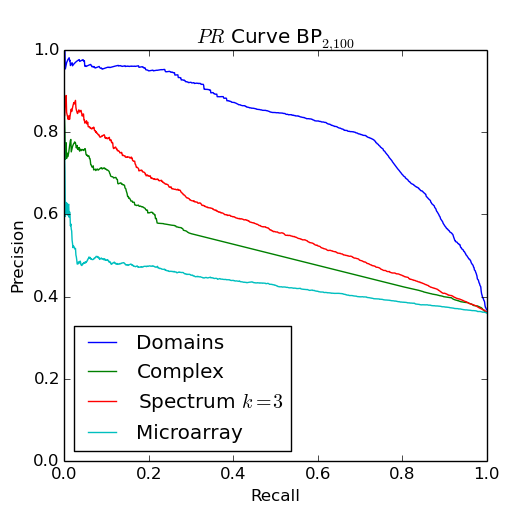}
	\end{subfigure}
    \begin{subfigure}[]{0.45\textwidth}
    	\centering
		\includegraphics[width=.9\textwidth]{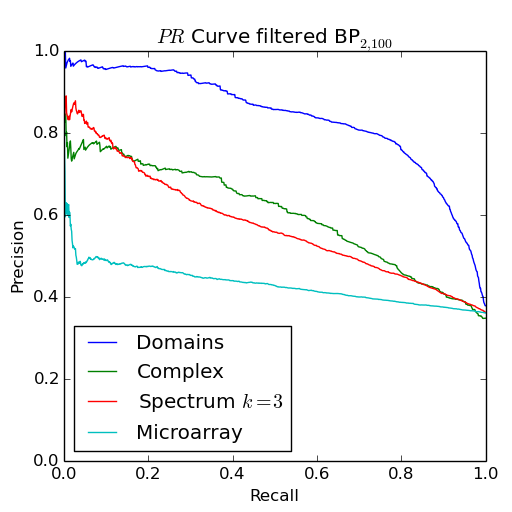}
	\end{subfigure}
\vspace{-5pt}
\myCaption{Precision-Recall Curves of the kernel function for Molecular Function $_{2,100}$}{In the two plots are reported the PR curves obtained on MF$_{2,100}$ without rules. The right plot shows the effects of the filtering on the curves.}
\label{fig:comp_MF}
\end{figure}

Figure \ref{fig:comp_MF} shows the comparison of the precision-recall curves obtained with the different kernels with on without the filtering of the undecided values. The graphs confirms the observation made on the result tables. It is clear that in both conditions the domains kernel outperforms the others, maintaining a high precision (over $80\%$) until almost $80\%$ o recall. Moreover we can notice that, despite being just marginally related with the molecular function hierarchy, the complex kernel positively responds to the filtering also in this situation.

The analysis made on the results with our kernels confirms the trends obtained by most of the systems found in literature \cite{res1,res2}, in which the prediction of molecular functions results to be an easier task if compared to the one of biological processes. A partial explanation this result lies into the consideration that molecular functions are intrinsic properties of the proteins and just marginally depend on the surrounding environment. Is therefore easier to detect important feature ``in'' the protein, which are also more abundant and precise thanks to the new sequencing techniques.

\subsection*{Ontology Consistency constraints}
\begin{table}[H]
\centering
\begin{tabulary}{\textwidth}{lCCC}
\toprule
               & $\mathcal{C}_\mathcal{E}$ & $\mathcal{C}_{\textrm{OC}}$ & \% increment \\ \midrule
Complexes      & 0.139          & 0.141          & 0.862          \\
Domains        & 0.249          & 0.250          & 0.104          \\
\rowcolor[HTML]{EFEFEF} 
Microarray     & \textbf{0.266} & \textbf{0.407} & 52.96          \\
Spectrum $k=3$ & 0.081          & 0.170          & \textbf{110.8} \\ \bottomrule
\end{tabulary}
\myCaption{Consistency of the predictions with OC rules on MF $_{2,100}$}{The table shows the comparison of the consistency measure $\mathcal{C}$ (see Section \ref{sec:perf}) in the experiments with and without consistency rules. The last column is the percentage increase obtained with the rules.}
\label{tab:MF_con}
\end{table}
\begin{table}[t]
\centering
\begin{tabulary}{\textwidth}{clCCCCCC}
\toprule
Rules & Kernel & $P^\lambda_\mu$ & $R^\lambda_\mu$ & $F1^\lambda_\mu$ & $P^\lambda_M$ & $R^\lambda_M$ & $F1^\lambda_M$ \\ \midrule
 & Complexes & 0.415 & 0.884 & 0.565 & 0.404 & 0.869 & 0.527 \\
 & \cellcolor[HTML]{EFEFEF}Domains & \cellcolor[HTML]{EFEFEF}\textbf{0.700} & \cellcolor[HTML]{EFEFEF}\textbf{0.805} & \cellcolor[HTML]{EFEFEF}\textbf{0.749} & \cellcolor[HTML]{EFEFEF}\textbf{0.650} & \cellcolor[HTML]{EFEFEF}\textbf{0.783} & \cellcolor[HTML]{EFEFEF}\textbf{0.701} \\
\multirow{-2}{*}{$\mathcal{E}$} & Microarray & 0.437 & 0.565 & 0.493 & 0.418 & 0.545 & 0.452 \\
 & Spectrum $k=3$ & 0.549 & 0.798 & 0.650 & 0.491 & 0.677 & 0.555 \\ \midrule
 & Complexes & 0.415 & 0.885 & 0.565 & 0.404 & 0.870 & 0.527 \\
 & \cellcolor[HTML]{EFEFEF}Domains & \cellcolor[HTML]{EFEFEF}\textbf{0.701} & \cellcolor[HTML]{EFEFEF}\textbf{0.806} & \cellcolor[HTML]{EFEFEF}\textbf{0.750} & \cellcolor[HTML]{EFEFEF}\textbf{0.650} & \cellcolor[HTML]{EFEFEF}\textbf{0.784} & \cellcolor[HTML]{EFEFEF}\textbf{0.702} \\
\multirow{-2}{*}{OC} & Microarray & 0.449 & 0.613 & 0.518 & 0.411 & 0.566 & 0.461 \\
 & Spectrum $k=3$ & 0.568 & 0.769 & 0.653 & 0.503 & 0.645 & 0.552 \\ \bottomrule
\end{tabulary}
\myCaption{Effects of the OC rules on Molecular Function $_{2,100}$}{}
\label{tab:MF_OC}
\end{table}

Table \ref{tab:MF_con} and Table \ref{tab:MF_OC} report respectively the effect of the OC constraints on the overall consistency of the prediction and on their quality. Similar consideration to the ones made for the biological process case can be applied also here. This confirms the hypothesis that the right choice of the meta-parameter is more related to the kernel rather than to the analyzed task. We can indeed notice that complexes and domains kernel have been just marginally affected by the constraints, on the other hand on microarray and spectrum ones the OC rules deeply impacted on the consistency measure. However, in this case, the improvement in the consistency of the predictions is not completely transferred to their quality. The microarray kernel, which has the greatest gains in the consistency metric (in absolute value), has just a $2.5\%$ increase in the  $F1^\lambda_\mu$ measure.

\begin{table}[t]
\centering
\begin{tabulary}{\textwidth}{CCCCCCCCC}
\toprule
Rules & $\lambda_C$ & $P^\lambda_\mu$ & $R^\lambda_\mu$ & $F1^\lambda_\mu$ & $P^\lambda_M$ & $R^\lambda_M$ & $F1^\lambda_M$ & $\mathcal{C}$ \\ \midrule
$\mathcal{E}$ &  & 0.700 & 0.805 & 0.749 & 0.650 & 0.783 & 0.701 & 0.249 \\\midrule
   & $10^1$ & 0.698 & 0.828 & 0.757 & 0.646 & 0.798 & 0.706 & 0.358 \\
OC & $10^3$ & 0.590 & 0.904 & 0.714 & 0.565 & 0.869 & 0.667 & 0.771 \\
   & $10^5$ & 0.444 & 0.968 & 0.609 & 0.423 & 0.929 & 0.566 & 0.995 \\ \bottomrule
\end{tabulary}
\myCaption{Effects of $\lambda_C$ on the OC rules}{The table shows the effect of the constraint strictness on the prediction of MF$_{2,100}$ with the domains kernel.}
\label{tab:lambda_C}
\end{table}

The examples shown in Table \ref{tab:lambda_C} make the relation between the strictness of the constrains and the quality of the performance clearer. It is indeed easy to notice that the increase of the consistency measure do not directly implies a benefit in term of performance. The last row shows an experiment made on purpose to demonstrate that, despite being completely satisfied in the training set, hard constraints (the value of $\lambda_C$ is so high to dominate in the optimization process) may strongly worsen the quality of prediction.

\newpage
%\subsection*{Protein-Protein Interaction constraints}
%bf

\section{Joint prediction of Biological Process and Molecular Function}

Part of the goals of this work was to analyze the effect of trans-hierarchy relationships on the final prediction of protein features. In this section we will therefore analyze the results obtained on the joint prediction of Molecular Function and Biological Process. Being the number of analyzed rules proportional to the number of predicates, the prediction of both hierarchies becomes soon computationally very hard. We decided therefore to limit the search to the first two levels of Gene Ontology with a count threshold $c=100$. Unfortunately with this restrictions the only GO cross-hierarchy relation taken into account is:
$$
	\overset{\small \bf \textrm{MF}}{\texttt{CATALITIC-ACTIVITY}}~~ \overset{part~of}{\xrightarrow{\hspace*{2cm}}}~~ \overset{\small \bf \textrm{BP}}{\texttt{METABOLIC-PROCESS}}
$$
This scarceness, how will be shown, make the information flow between BP and MF just marginally effective.

In Table \ref{tab:MF_OC} are reported the results obtained in these experiments with and without constraints. It can be notice that the kernels reflect the observation made for the disjointed predictions. It is important to notice that the number of predicted predicates is larger in the Biological Process hierarchy making the contribute of Molecular Function slightly less significant.

The domains kernel obtains the bests $F1$ scores, both in the constrained and unconstrained cases, confirming its leading position in the unfiltered condition. Moreover, we can appreciate the effect of OC rules on the spectrum kernel, which gains almost a $5\%$  in the constrained case. This confirm the results highlighted in the previous sections, i.e. the kernel most affected by constraints are smother ones. 

The introduction of the DPPI rules proposes very interesting results. Besides confirming that valid constraints positively impact on the overall quality of predictions, it shows that rules and kernels can effectively work together and mix their strengths. This behavior in evident when compared the two most informative kernels, i.e the complexes and the domains ones, in the DPPI setting. It can be noticed that the complexes kernel is just minimally affected by this set of rules, on the other hand the domains one has a very positive improvement. We know that the complexes kernel is based on the protein complexes interaction network that is a subset of the PPI. Therefore the increment of information in this case is marginal. Differently the domains kernel has no internally information on protein protein interaction. Hence the constraints inject a considerable quantity of new knowledge that is translated into an improvement of the overall predictions quality.

\begin{table}[t]
\centering
\begin{tabulary}{\textwidth}{clCCCCCC}
\toprule
Rules & Kernel & $P^\lambda_\mu$ & $R^\lambda_\mu$ & $F1^\lambda_\mu$ & $P^\lambda_M$ & $R^\lambda_M$ & $F1^\lambda_M$ \\ \midrule
 & Complexes & 0.452 & \textbf{0.922} & 0.606 & 0.431 & \textbf{0.906} & 0.555 \\
 & \cellcolor[HTML]{EFEFEF}Domains & \cellcolor[HTML]{EFEFEF}\textbf{0.622} & \cellcolor[HTML]{EFEFEF}0.724 & \cellcolor[HTML]{EFEFEF}\textbf{0.669} & \cellcolor[HTML]{EFEFEF}\textbf{0.571} & \cellcolor[HTML]{EFEFEF}0.702 & \cellcolor[HTML]{EFEFEF}\textbf{0.613} \\
\multirow{-2}{*}{$\mathcal{E}$} & Microarray & 0.458 & 0.568 & 0.507 & 0.446 & 0.571 & 0.467 \\
 & Spectrum $k=3$ & 0.568 & 0.616 & 0.591 & 0.499 & 0.530 & 0.496 \\ \midrule
 & Complexes & 0.452 & \textbf{0.922} & 0.606 & 0.431 & \textbf{0.906} & 0.555 \\
 & \cellcolor[HTML]{EFEFEF}Domains & \cellcolor[HTML]{EFEFEF}\textbf{0.622} & \cellcolor[HTML]{EFEFEF}0.724 & \cellcolor[HTML]{EFEFEF}\textbf{0.669} & \cellcolor[HTML]{EFEFEF}\textbf{0.571} & \cellcolor[HTML]{EFEFEF}0.701 & \cellcolor[HTML]{EFEFEF}\textbf{0.613} \\
\multirow{-2}{*}{OC} & Microarray & 0.475 & 0.608 & 0.533 & 0.444 & 0.584 & 0.479 \\
 & Spectrum $k=3$ & 0.595 & 0.662 & 0.627 & 0.505 & 0.541 & 0.510 \\ \midrule
 & Complexes & 0.453 & \textbf{0.922} & 0.607 & 0.432 & \textbf{0.905} & 0.555 \\
 & \cellcolor[HTML]{EFEFEF}Domains & \cellcolor[HTML]{EFEFEF}\textbf{0.667} & \cellcolor[HTML]{EFEFEF}0.750 & \cellcolor[HTML]{EFEFEF}\textbf{0.706} & \cellcolor[HTML]{EFEFEF}\textbf{0.592} & \cellcolor[HTML]{EFEFEF}0.694 & \cellcolor[HTML]{EFEFEF}\textbf{0.633} \\
\multirow{-2}{*}{DPPI} & Microarray & 0.467 & 0.570 & 0.513 & 0.440 & 0.551 & 0.464 \\
 & Spectrum $k=3$ & 0.634 & 0.595 & 0.614 & 0.525 & 0.463 & 0.481 \\ \bottomrule
\end{tabulary}
\myCaption{Results for native joint prediction of BP $_{2,100}$ and MF $_{2,100}$}{}
\label{tab:MF_BP_OC}
\end{table}

\chapter{Conclusion}

\label{ch:conclusions}

%\section{Summary of Thesis Achievements}
The aim of this work was to exploit a state-of-the-art statistical relational learning framework to predict protein features. We focused our analysis on the prediction of the annotations belonging to the first levels of the biological process and molecular functions hierarchies of Gene Ontology, which is a structured vocabulary of the definition of genes products (in our case proteins). The framework employed in this work is called Semantic Based Regularization and allowed us to define first order logic rules that can constrain the typical kernel machine learning process.

The main efforts in this work have been oriented to the designing of effective but valid constraints, that could exploits different biological sources to refine the quality of prediction, and to the implementation of four kernel functions that covers various aspects of proteins nature. The whole setup has been then tested on more then 1500 proteins of the \emph{S. cerevisiae}, the baker yeast.

From the experimental results we were able to appreciate the effects of the constraints on the predictions. Despite the suboptimal choice of the meta parameters, we noticed good improvements with the ontology consistency rules, in particular when applied in combination with `smoother' kernel. Differently, with kernels where information is very sharp and defines, like the domains and complexes ones, are needed higher values of the $\lambda_C$ parameter to impact on the results. However, too strict constraints will considerably worsen the prediction. Moreover is important to verify the validity of the rules, because the injection of misleading information will clearly lead to a negative impact on the performance.

The prediction of the biological processes related to a protein confirms the trend found in the literature  and was a complex task. Indeed the final prediction with the best kernel (domains kernel) were just over $60\%$ of the $F1_\mu$ measure. However the strong unbalance of the complex kernel results led us to notice that for several proteins the system was not able to take a decision and the predictions converged in the threshold. This behavior can be attributed to the sparseness of this kernel and, being these proteins easily recognizable, they can be removed from the analysis. In this situation the complexes kernel shows its correlation to the biological processes and achieves an $F1_\mu$ measure around the $80\%$. The obtained results are fully admissible from a biological point of view. It is indeed true that domains are just marginally related to biological processes; two proteins can share very similar domains and, despite this, being part of two completely different biological process. On the other hand, two proteins part of the same protein complex are very likely to be part of the same biological process.

The prediction of molecular functions gave us good results. We have been able to predict them with more of the $70\%$ of $F1_\mu$ measure in all the analyzed levels thanks to the domains kernel. This confirms the hypothesis that protein domains are highly related to their molecular function. It was interesting to notice that, on the filtered results, the quality of the results (always over the $80\%$ of $F1_\mu$) increased with the depth of the levels. Our conclusion is that domains are so specific for the MF annotations, that high level nodes of Gene Ontology are too general and mislead some predictions. In addition to the domains kernel, it is interesting to notice the behavior of the spectrum kernel. Indeed, despite its simplicity and the little information used (just the amino acids sequence), it obtains decent results. This leads to the conclusion that the amino acid sequence plays a crucial role in the prediction of molecular function, much more than for the biological process. It is indeed true that molecular functions strongly depends on the structure and composition of a protein and are just marginally influenced by the surrounding environment. Therefore the information needed to predict them can be found directly `in' the protein, which thanks to the improvements in the sequencing techniques is always more abundant and precise. On the other hand biological processes strongly depends on the environment surrounding the protein. It is indeed the conjunction of subcellular localization, molecular functions and interactions, and the moment in the cell cycle to determine the biological processes a protein is part of. This make the task much more complex. Moreover the information are mostly located `outside' the protein and therefore harder to retrieve.

Concluding, we obtained good results with the molecular function annotations, but more work has to be done in the prediction biological processes in order to achieve the reliability required to offer clear target that biologist could exploit to focus their laboratory researches. Nonetheless, this work shows the effectiveness of combining kernel functions and logical constraints in the prediction of biological features. Indeed, our results represent a good starting point for further works in this direction.

%\section{Future Work}

%profile kernels

\appendix
\begin{appendices}
\chapter*{Protein-Protein interaction}

\begin{table}[htp]
\centering
\begin{tabular}{@{}llccc@{}}
\toprule
Level              & Predicate                           & POS PP & TOT PP & RATIO \\ \midrule
0                  & \tt MOLECULAR-FUNCTION              & 2688         & 2688         & 1.000 \\\midrule
\multirow{3}{*}{1} & \tt BINDING                         & 1848         & 2458         & 0.752 \\
                   & \tt CATALYTIC-ACTIVITY              & 693          & 1647         & 0.421 \\
                   & \tt ENZYME-REGULATOR-ACTIVITY       & 62           & 451          & 0.137 \\ \midrule
\multirow{8}{*}{2} & \tt ORGANIC-CYCLIC-COMPOUND-BINDING & 937          & 1815         & 0.516 \\
                   & \tt HETEROCYCLIC-COMPOUND-BINDING   & 936          & 1814         & 0.516 \\
                   & \tt ION-BINDING                     & 657          & 1593         & 0.412 \\
                   & \tt PROTEIN-BINDING                 & 448          & 1267         & 0.354 \\
                   & \tt TRANSFERASE-ACTIVITY            & 199          & 631          & 0.315 \\
                   & \tt SMALL-MOLECULE-BINDING          & 346          & 1212         & 0.285 \\
                   & \tt HYDROLASE-ACTIVITY              & 251          & 929          & 0.270 \\
                   & \tt CARBOHYDRATE-DERIVATIVE-BINDING & 282          & 1061         & 0.266 \\  \bottomrule
\end{tabular}
\myCaption{PPI predicate sharing for Molecular Function$_{2,100}$}{In this table are reported the values of the predicate coverage for the protein in the interaction dataset. The values has been calculated on the real predicate and proteins for the MF$_{2,100}$. \emph{TOP PP} indicates the number of protein pair for which at least one of the two has the corresponding predicate. \emph{POS PP} instead in the number of protein pairs having both the predicate. the last column is the ratio between the previous two.}
\label{tab:pp1}
\end{table}

\begin{table}[htp]
\centering
\begin{tabular}{@{}llccc@{}}
\toprule
Level               & Predicate                                             & POS PP & TOT PP & RATIO \\ \midrule 
0                   & \tt BIOLOGICAL-PROCESS                          & 3125         & 3125         & 1.000 \\\midrule
\multirow{7}{*}{1}  & \tt CELLULAR-PROCESS                            & 2681         & 3036         & 0.883 \\
                    &\tt SINGLE-ORGANISM-PROCESS                       & 1930         & 2615         & 0.738 \\
                    & \tt METABOLIC-PROCESS                             & 1938         & 2661         & 0.728 \\
                    & \tt BIOLOGICAL-REGULATION                         & 1233         & 2066         & 0.597 \\
                    & \tt LOCALIZATION                                  & 732          & 1354         & 0.541 \\
                    & \tt CELLULAR-COMPONENT-ORGANIZATION... & 1162         & 2195         & 0.529 \\
                    & \tt RESPONSE-TO-STIMULUS                          & 510          & 1312         & 0.389 \\\midrule
\multirow{22}{*}{2} & \tt CELLULAR-METABOLIC-PROCESS                    & 1866         & 2607         & 0.716 \\
                    & \tt ORGANIC-SUBSTANCE-METABOLIC-PROCESS           & 1786         & 2517         & 0.710 \\
                    & \tt PRIMARY-METABOLIC-PROCESS                     & 1738         & 2471         & 0.703 \\
                    & \tt SINGLE-ORGANISM-CELLULAR-PROCESS              & 1592         & 2437         & 0.653 \\
                    & \tt REGULATION-OF-BIOLOGICAL-PROCESS              & 1129         & 1963         & 0.575 \\
                    & \tt NITROGEN-COMPOUND-METABOLIC-PROCESS           & 1064         & 1882         & 0.565 \\
                    & \tt ESTABLISHMENT-OF-LOCALIZATION                 & 654          & 1219         & 0.537 \\
                    & \tt CELLULAR-COMPONENT-ORGANIZATION               & 1105         & 2160         & 0.512 \\
                    & \tt BIOSYNTHETIC-PROCESS                          & 533          & 1046         & 0.510 \\
                    & \tt SINGLE-ORGANISM-LOCALIZATION                  & 540          & 1090         & 0.495 \\
                    & \tt SINGLE-ORGANISM-TRANSPORT                     & 486          & 1010         & 0.481 \\
                    & \tt MACROMOLECULE-LOCALIZATION                    & 425          & 895          & 0.475 \\
                    & \tt SINGLE-ORGANISM-METABOLIC-PROCESS             & 736          & 1657         & 0.444 \\
                    & \tt ORGANELLE-ORGANIZATION                        & 623          & 1545         & 0.403 \\
                    & \tt CATABOLIC-PROCESS                            & 496          & 1269         & 0.391 \\
                    & \tt CELLULAR-RESPONSE-TO-STIMULUS                 & 436          & 1188         & 0.367 \\
                    & \tt RESPONSE-TO-STRESS                          & 310          & 910          & 0.341 \\
                    & \tt CELLULAR-LOCALIZATION                        & 233          & 689          & 0.338 \\
                    & \tt SINGLE-ORGANISM-MEMBRANE-ORGANIZATION         & 125          & 403          & 0.310 \\
                    & \tt REGULATION-OF-BIOLOGICAL-QUALITY              & 151          & 510          & 0.296 \\
                    & \tt RESPONSE-TO-CHEMICAL                          & 78           & 387          & 0.202 \\
                    & \tt REGULATION-OF-MOLECULAR-FUNCTION             & 137          & 695          & 0.197 \\ \bottomrule
\end{tabular}
\myCaption{PPI predicate sharing for Biological Process$_{2,100}$}{In this table are reported the values of the predicate coverage for the protein in the interaction dataset. The values has been calculated on the real predicate and proteins for the BP$_{2,100}$. \emph{TOP PP} indicates the number of protein pair for which at least one of the two has the corresponding predicate. \emph{POS PP} instead in the number of protein pairs having both the predicate. the last column is the ratio between the previous two.}
\label{tab:pp2}
\end{table}

\chapter*{Kernel sparseness}
\label{ap:kernel}
\begin{figure}[H]
	\vspace{-1cm}
    \centering
    \begin{subfigure}[]{0.75\textwidth}
        \centering
        \includegraphics[width=\textwidth]{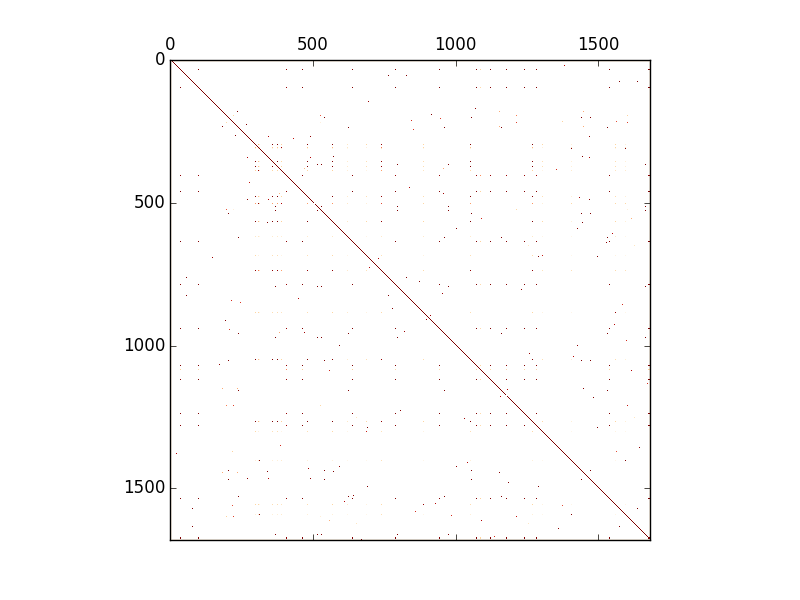}
        \caption{Complex-based kernel}
        \label{fig:complex}
    \end{subfigure}
    \begin{subfigure}[]{0.75\textwidth}
        \centering
        \includegraphics[width=\textwidth]{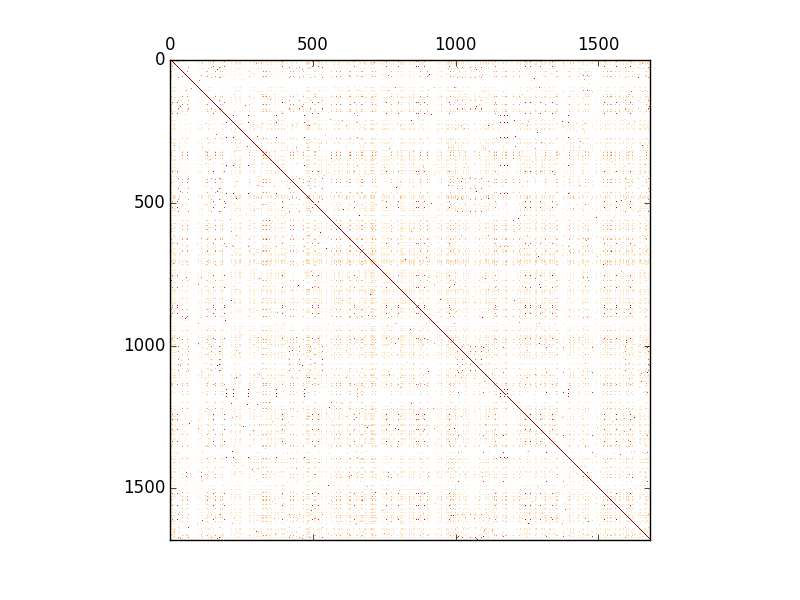}
        \caption{Domain-based kernel}
        \label{fig:domains}
    \end{subfigure}
    \label{fig:three graphs}
    \phantomcaption
\end{figure}

\begin{figure}
	%\ContinuedFloat
    \centering
    \begin{subfigure}[]{0.75\textwidth}
        \centering
        \includegraphics[width=\textwidth]{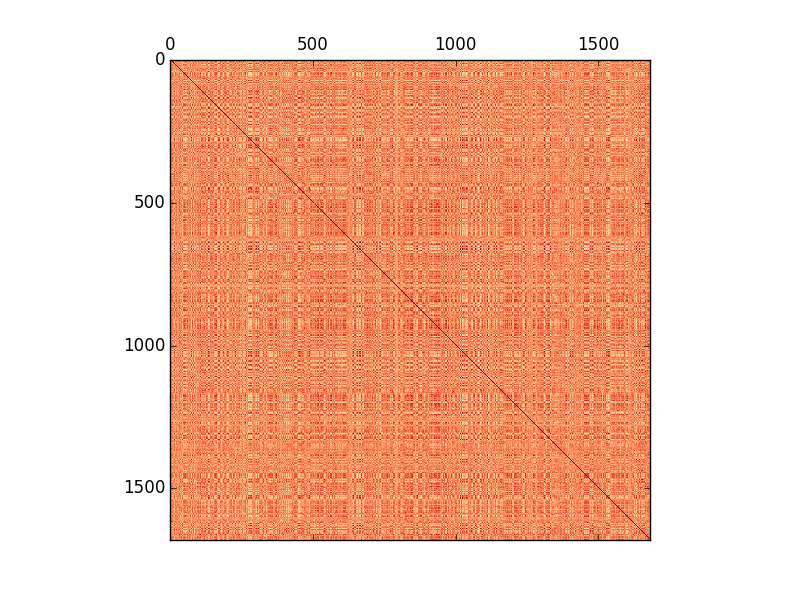}
        \caption{Microarray kernel}
        \label{fig:microarray}
    \end{subfigure}
    \begin{subfigure}[]{0.75\textwidth}
        \centering
        \includegraphics[width=\textwidth]{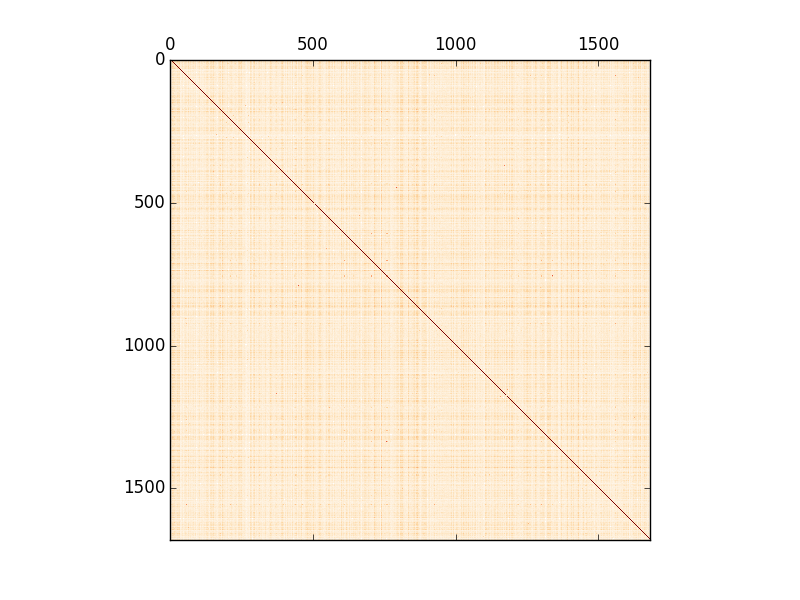}
        \caption{Spectrum kernel $k=3$}
        \label{fig:spectrum}
    \end{subfigure}
    \myCaption{Kernel matrix heatmaps}{The proposed heatmaps refer to the kernel matrix we implemented. The intensity of the color for each coordinate in the plots is proportional to the kernel value for that pair of points. }
    \label{fig:heatmaps}
\end{figure}

\chapter*{Curve averaging algorithms results}

\begin{figure}[H]
\centering
\includegraphics[width=1\textwidth]{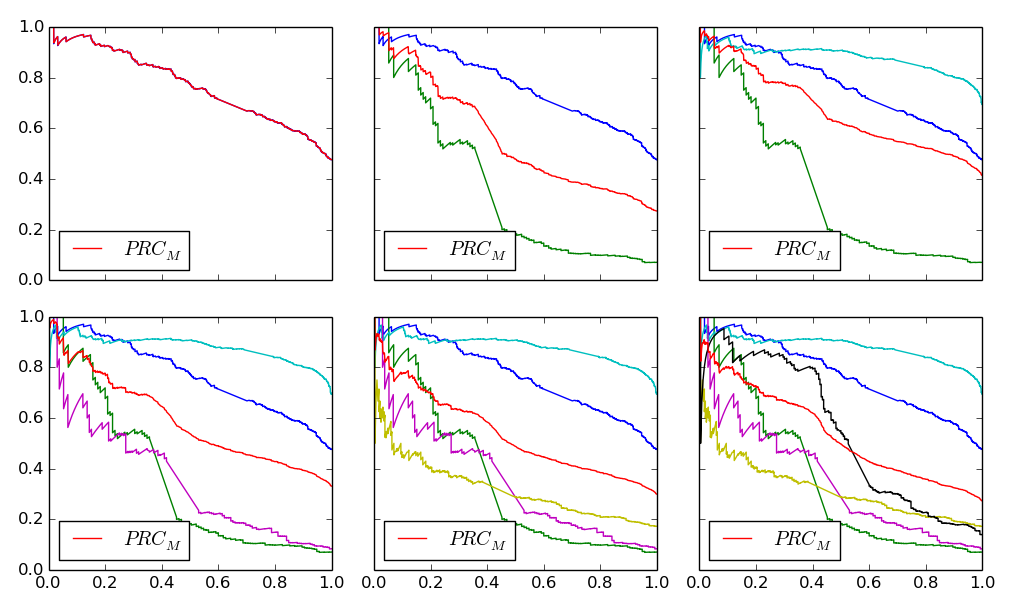}
\myCaption{\emph{Macro} averaged Precision-Recall Curve Example}{}
\label{fig:curves}
\end{figure}

\chapter*{Results Tables}

\begin{table}[htp]
\begin{tabular}{@{}lccccccc@{}}
\toprule
Kernel                          & Rules      & Precision$_\mu$ & Recall$_\mu$ & F1$_\mu$ & Precision$_M$ & Recall$_M$ & F1$_M$ \\ \midrule
\multirow{2}{*}{Complex}        & $\epsilon$ & 0.415         & 0.884      & 0.565  & 0.404         & 0.869      & 0.527  \\
                                & OC         & 0.415         & 0.885      & 0.565  & 0.404         & 0.870      & 0.527  \\\midrule
                             %  & PP1        & 0.431         & 0.874      & 0.577  & 0.433         & 0.862      & 0.547  \\
                             %  & PP2        & 0.415         & 0.884      & 0.565  & 0.404         & 0.869      & 0.527  \\
\multirow{2}{*}{Domains}        & $\epsilon$ & 0.700         & 0.805      & 0.749  & 0.650         & 0.783      & 0.701  \\
                                & OC         & 0.701         & 0.806      & 0.750  & 0.650         & 0.784      & 0.702  \\\midrule
                              % & PP1        & 0.705         & 0.803      & 0.750  & 0.658         & 0.783      & 0.707  \\
                              % & PP2        & 0.700         & 0.805      & 0.749  & 0.650         & 0.783      & 0.701  \\
\multirow{2}{*}{Microarray}     & $\epsilon$ & 0.437         & 0.565      & 0.493  & 0.418         & 0.545      & 0.452  \\
                                & OC         & 0.449         & 0.613      & 0.518  & 0.411         & 0.566      & 0.461  \\\midrule
                              % & PP1        & 0.466         & 0.605      & 0.526  & 0.441         & 0.572      & 0.481  \\
                              % & PP2        & 0.443         & 0.588      & 0.505  & 0.413         & 0.554      & 0.456  \\
\multirow{2}{*}{Spectrum $k=3$} & $\epsilon$ & 0.549         & 0.798      & 0.650  & 0.491         & 0.677      & 0.555  \\
                                & OC         & 0.568         & 0.769      & 0.653  & 0.503         & 0.645      & 0.552  \\\bottomrule
                               %& PP1        & 0.582         & 0.766      & 0.662  & 0.525         & 0.654      & 0.569  \\
                               %& PP2        & 0.565         & 0.771      & 0.652  & 0.504         & 0.649      & 0.553  \\ 
\end{tabular}
\myCaption{Results Molecular Function$_{2,100}$}{Results obtained with the four kernels on the Molecular Function with $level\_threshold=2$ and $count\_threshold=100$. The notation \emph{metric}$_\mu$ stays for \emph{micro average} and  \emph{metric}$_M$ stays for \emph{macro average}. The definition of the rules are reported in the Methods chapter (Section \ref{sec:rules}).}
\end{table}

\begin{table}[htp]
\begin{tabular}{@{}lccccccc@{}}
\toprule
Kernel                          & Rules      & Precision$_\mu$ & Recall$_\mu$ & F1$_\mu$ & Precision$_M$ & Recall$_M$ & F1$_M$ \\ \midrule
\multirow{2}{*}{Complex}        & $\epsilon$ & 0.347         & 0.874      & 0.496  & 0.337         & 0.861      & 0.464  \\
                                & OC         & 0.347         & 0.874      & 0.496  & 0.337         & 0.862      & 0.464  \\\midrule
                              % & PP1        & 0.358         & 0.868      & 0.507  & 0.356         & 0.858      & 0.478  \\
                              % & PP2        & 0.347         & 0.874      & 0.496  & 0.337         & 0.862      & 0.464  \\
\multirow{2}{*}{Domains}        & $\epsilon$ & 0.665         & 0.829      & 0.738  & 0.625         & 0.827      & 0.704  \\
                                & OC         & 0.665         & 0.829      & 0.738  & 0.625         & 0.826      & 0.704  \\\midrule
                              % & PP1        & 0.669         & 0.826      & 0.74   & 0.632         & 0.824      & 0.707  \\
                              % & PP2        & 0.665         & 0.829      & 0.738  & 0.625         & 0.826      & 0.704  \\
\multirow{2}{*}{Microarray}     & $\epsilon$ & 0.363         & 0.571      & 0.444  & 0.353         & 0.561      & 0.412  \\
                                & OC         & 0.363         & 0.572      & 0.444  & 0.352         & 0.561      & 0.412  \\\midrule
                              % & PP1        & 0.380         & 0.584      & 0.461  & 0.372         & 0.572      & 0.429  \\
                              % & PP2        & 0.363         & 0.571      & 0.444  & 0.353         & 0.561      & 0.412  \\
                                & $\epsilon$ & 0.520         & 0.730      & 0.608  & 0.477         & 0.624      & 0.525  \\
\multirow{-2}{*}{Spectrum $k=3$}& OC         & 0.520         & 0.730      & 0.607  & 0.478         & 0.623      & 0.525  \\\bottomrule
                             %  & PP1        & 0.531         & 0.733      & 0.616  & 0.491         & 0.630      & 0.536  \\
                             %  & PP2        & 0.520         & 0.730      & 0.607  & 0.478         & 0.623      & 0.524  \\ 
\end{tabular}
\myCaption{Results Molecular Function$_{3,100}$}{Results obtained with the four kernels on the Molecular Function with $level\_threshold=3$ and $count\_threshold=100$. The notation \emph{metric}$_\mu$ stays for \emph{micro average} and  \emph{metric}$_M$ stays for \emph{macro average}. The definition of the rules are reported in the Methods chapter (Section \ref{sec:rules}).}
\end{table}

\begin{table}[htp]
\begin{tabular}{@{}lccccccc@{}}
\toprule
Kernel                          & Rules      & Precision$_\mu$ & Recall$_\mu$ & F1$_\mu$ & Precision$_M$ & Recall$_M$ & F1$_M$ \\ \midrule
\multirow{2}{*}{Complex}        & $\epsilon$ & 0.305         & 0.869      & 0.452  & 0.297         & 0.861      & 0.423  \\
                                & OC         & 0.305         & 0.869      & 0.452  & 0.297         & 0.861      & 0.423  \\\midrule
\multirow{2}{*}{Domains}        & $\epsilon$ & 0.641         & 0.841      & 0.727  & 0.604         & 0.839      & 0.695  \\
                                & OC         & 0.641         & 0.841      & 0.727  & 0.604         & 0.839      & 0.695  \\\midrule
                               % & PP1        & 0.645         & 0.837      & 0.729  & 0.608         & 0.837      & 0.697  \\
                               % & PP2        & 0.641         & 0.841      & 0.727  & 0.604         & 0.839      & 0.695  \\
\multirow{2}{*}{Microarray}     & $\epsilon$ & 0.324         & 0.562      & 0.411  & 0.314         & 0.560      & 0.384  \\
                                & OC         & 0.324         & 0.563      & 0.411  & 0.314         & 0.560      & 0.384  \\\midrule
                               % & PP1        & 0.336         & 0.572      & 0.423  & 0.328         & 0.567      & 0.396  \\
                               % & PP2        & 0.324         & 0.562      & 0.411  & 0.314         & 0.560      & 0.384  \\
                                & $\epsilon$ & 0.496         & 0.696      & 0.579  & 0.463         & 0.595      & 0.504  \\
\multirow{-2}{*}{Spectrum $k=3$} & OC         & 0.497         & 0.696      & 0.579  & 0.463         & 0.593      & 0.504  \\\bottomrule
                               % & PP1        & 0.506         & 0.699      & 0.587  & 0.473         & 0.599      & 0.512  \\
                               % & PP2        & 0.497         & 0.695      & 0.579  & 0.463         & 0.593      & 0.503  \\ 
\end{tabular}
\myCaption{Results Molecular Function$_{4,100}$}{Results obtained with the four kernels on the Molecular Function with $level\_threshold=4$ and $count\_threshold=100$. The notation \emph{metric}$_\mu$ stays for \emph{micro average} and  \emph{metric}$_M$ stays for \emph{macro average}. The definition of the rules are reported in the Methods chapter (Section \ref{sec:rules}).}
\end{table}

\begin{table}[htp]
\begin{tabular}{@{}lccccccc@{}}
\toprule
Kernel                          & Rules      & Precision$_m$ & Recall$_m$ & F1$_m$ & Precision$_M$ & Recall$_M$ & F1$_M$ \\ \midrule
                                & $\epsilon$ & 0.460         & 0.932      & 0.616  & 0.435         & 0.920      & 0.557  \\
\multirow{2}{*}{Complex}        & OC         & 0.460         & 0.932      & 0.616  & 0.435         & 0.920      & 0.557  \\
                              % & PP1        & 0.465         & 0.926      & 0.619  & 0.446         & 0.915      & 0.564  \\
                                & PPI        & 0.460         & 0.932      & 0.616  & 0.435         & 0.920      & 0.557  \\
                              % & DPP1       & 0.467         & 0.928      & 0.621  & 0.447         & 0.913      & 0.565  \\
                                & DPPI       & 0.461         & 0.934      & 0.618  & 0.436         & 0.918      & 0.558  \\\midrule
                                & $\epsilon$ & 0.575         & 0.682      & 0.624  & 0.520         & 0.656      & 0.558  \\
\multirow{2}{*}{Domains}        & OC         & 0.575         & 0.682      & 0.624  & 0.521         & 0.656      & 0.559  \\
                               %& PP1        & 0.581         & 0.685      & 0.629  & 0.529         & 0.660      & 0.565  \\
                                & PPI        & 0.575         & 0.682      & 0.624  & 0.521         & 0.656      & 0.559  \\
                              % & DPP1       & 0.630         & 0.700      & 0.663  & 0.551         & 0.641      & 0.583  \\
                                & DPPI       & 0.625         & 0.699      & 0.66   & 0.544         & 0.638      & 0.578  \\\midrule
                                & $\epsilon$ & 0.453         & 0.568      & 0.504  & 0.440         & 0.576      & 0.460  \\
                                & OC         & 0.475         & 0.621      & 0.538  & 0.439         & 0.594      & 0.476  \\
\multirow{2}{*}{Microarray}    %& PP1        & 0.474         & 0.602      & 0.531  & 0.451         & 0.590      & 0.480  \\
                                & PPI        & 0.467         & 0.597      & 0.524  & 0.440         & 0.584      & 0.470  \\
                               %& DPP1       & 0.469         & 0.586      & 0.521  & 0.442         & 0.565      & 0.467  \\
                                & DPPI       & 0.447         & 0.536      & 0.487  & 0.415         & 0.512      & 0.431  \\\midrule
                                & $\epsilon$ & 0.569         & 0.536      & 0.552  & 0.496         & 0.458      & 0.457  \\
\multirow{2}{*}{Spectrum $k=3$} & OC         & 0.601         & 0.621      & 0.611  & 0.496         & 0.491      & 0.482  \\
                                %& PP1        & 0.602         & 0.605      & 0.604  & 0.506         & 0.491      & 0.487  \\
                                & PPI        & 0.596         & 0.599      & 0.598  & 0.497         & 0.481      & 0.477  \\
                               % & DPP1       & 0.627         & 0.580      & 0.603  & 0.511         & 0.445      & 0.464  \\
                                & DPPI       & 0.624         & 0.576      & 0.599  & 0.503         & 0.434      & 0.454  \\ \bottomrule
\end{tabular}
\myCaption{Results Biological Process$_{2,100}$}{Results obtained with the four kernels on the Biological Process with $level\_threshold=2$ and $count\_threshold=100$. The notation \emph{metric}$_m$ stays for \emph{micro average} and  \emph{metric}$_M$ stays for \emph{macro average}. The definition of the rules are reported in the Methods chapter (Section \ref{sec:rules}). The experiments conducted on the last sets of rules (DPPI) have been executed on a newer version of SBRS that supports constraints in \emph{DNF} and uses a different t-norm. Therefore, the results of these experiments are non completely comparable withe the others.}
\end{table}

\begin{table}[htp]
\begin{tabular}{@{}lccccccc@{}}
\toprule
Kernel                          & Rules      & Precision$_m$ & Recall$_m$ & F1$_m$ & Precision$_M$ & Recall$_M$ & F1$_M$ \\ \midrule
                                & $\epsilon$ & 0.352         & 0.928      & 0.510  & 0.332         & 0.914      & 0.458  \\
\multirow{2}{*}{Complex}        & OC         & 0.352         & 0.928      & 0.510  & 0.332         & 0.914      & 0.458  \\
                               %& PP1        & 0.355         & 0.924      & 0.513  & 0.337         & 0.912      & 0.462  \\
                                & PPI        & 0.352         & 0.928      & 0.510  & 0.332         & 0.914      & 0.458  \\
                              % & DPP1       & 0.356         & 0.924      & 0.514  & 0.337         & 0.909      & 0.462  \\
                                & DPPI       & 0.353         & 0.927      & 0.511  & 0.332         & 0.911      & 0.458  \\\midrule
                                & $\epsilon$ & 0.493         & 0.705      & 0.580  & 0.442         & 0.681      & 0.517  \\
                                & OC         & 0.493         & 0.704      & 0.580  & 0.442         & 0.681      & 0.517  \\
\multirow{-2}{*}{Domains}      % & PP1        & 0.497         & 0.706      & 0.583  & 0.446         & 0.682      & 0.520  \\
                                & PPI        & 0.493         & 0.704      & 0.580  & 0.442         & 0.681      & 0.517  \\
                               %& DPP1       & 0.547         & 0.687      & 0.609  & 0.469         & 0.628      & 0.528  \\
                                & DPPI       & 0.543         & 0.685      & 0.606  & 0.466         & 0.627      & 0.525  \\\midrule
                                & $\epsilon$ & 0.361         & 0.575      & 0.443  & 0.341         & 0.565      & 0.396  \\
\multirow{2}{*}{Microarray}     & OC         & 0.364         & 0.582      & 0.448  & 0.341         & 0.567      & 0.398  \\
                              % & PP1        & 0.367         & 0.582      & 0.450  & 0.347         & 0.569      & 0.401  \\
                                & PPI        & 0.362         & 0.578      & 0.445  & 0.341         & 0.566      & 0.397  \\
                              % & DPP1       & 0.356         & 0.568      & 0.438  & 0.335         & 0.548      & 0.387  \\
                                & DPPI       & 0.356         & 0.582      & 0.442  & 0.328         & 0.554      & 0.387  \\\midrule
                                & $\epsilon$ & 0.498         & 0.487      & 0.492  & 0.421         & 0.399      & 0.396  \\
\multirow{2}{*}{Spectrum $k=3$} & OC         & 0.500         & 0.488      & 0.494  & 0.422         & 0.397      & 0.395  \\
                             %  & PP1        & 0.506         & 0.493      & 0.499  & 0.427         & 0.401      & 0.400  \\
                                & PPI        & 0.500         & 0.487      & 0.494  & 0.423         & 0.396      & 0.395  \\
                             %  & DPP1       & 0.564         & 0.487      & 0.523  & 0.445         & 0.348      & 0.378  \\
                                & DPPI       & 0.560         & 0.479      & 0.516  & 0.443         & 0.341      & 0.373  \\ \bottomrule
\end{tabular}
\myCaption{Results Biological Process$_{3,100}$}{Results obtained with the four kernels on the Biological Process with $level\_threshold=3$ and $count\_threshold=100$. The notation \emph{metric}$_\mu$ stays for \emph{micro average} and  \emph{metric}$_M$ stays for \emph{macro average}. The definition of the rules are reported in the Methods chapter (Section \ref{sec:rules}). The experiments conducted on the last sets of rules (DPPI) have been executed on a newer version of SBRS that supports constraints in \emph{DNF} and uses a different t-norm. Therefore, the results of these experiments are non completely comparable withe the others.}
\end{table}

\begin{table}[htp]
\begin{tabular}{@{}lccccccc@{}}
\toprule
Kernel                          & Rules      & Precision$_\mu$ & Recall$_\mu$ & F1$_\mu$ & Precision$_M$ & Recall$_M$ & F1$_M$ \\ \midrule
                                & $\epsilon$ & 0.452         & 0.922      & 0.606  & 0.431         & 0.906      & 0.555  \\
\multirow{-2}{*}{Complex}        & OC         & 0.452         & 0.922      & 0.606  & 0.431         & 0.906      & 0.555  \\\midrule
                               % & PP1        & 0.456         & 0.918      & 0.609  & 0.44          & 0.903      & 0.56   \\
                               % & PP2        & 0.452         & 0.922      & 0.606  & 0.431         & 0.906      & 0.555  \\
                                & $\epsilon$ & 0.622         & 0.724      & 0.669  & 0.571         & 0.702      & 0.613  \\
\multirow{-2}{*}{Domains}        & OC         & 0.622         & 0.724      & 0.669  & 0.571         & 0.701      & 0.613  \\\midrule
%                                & PP1        &               &            &        &               &            &        \\
%                                & PP2        &               &            &        &               &            &        \\
                                & $\epsilon$ & 0.458         & 0.568      & 0.507  & 0.446         & 0.571      & 0.467  \\
\multirow{-2}{*}{Microarray}     & OC         & 0.475         & 0.608      & 0.533  & 0.444         & 0.584      & 0.479  \\\midrule
                               % & PP1        & 0.474         & 0.593      & 0.527  & 0.453         & 0.581      & 0.481  \\
                               % & PP2        & 0.468         & 0.588      & 0.521  & 0.445         & 0.577      & 0.474  \\
                                & $\epsilon$ & 0.568         & 0.616      & 0.591  & 0.499         & 0.53       & 0.496  \\
\multirow{-2}{*}{Spectrum $k=3$} & OC         & 0.595         & 0.662      & 0.627  & 0.505         & 0.541      & 0.51   \\\bottomrule
                              %  & PP1        & 0.591         & 0.645      & 0.617  & 0.512         & 0.539      & 0.512  \\
                              %  & PP2        & 0.587         & 0.643      & 0.614  & 0.505         & 0.534      & 0.505  \\ 
\end{tabular}
\myCaption{Results joint prediction of Biological Process and Molecular Function$_{4,100}$}{Results obtained with the four kernels on joint prediction of Biological Process and Molecular Function with $level\_threshold=4$ and $count\_threshold=100$. The notation \emph{metric}$_\mu$ stays for \emph{micro average} and  \emph{metric}$_M$ stays for \emph{macro average}. The definition of the rules are reported in the Methods chapter (Section \ref{sec:rules}).}
\end{table}

\end{appendices}

\addcontentsline{toc}{chapter}{Bibliography}
\bibliography{bibliography}
\bibliographystyle{plain}

\end{document}